%% file: alive.tex
\documentclass[acmtog]{acmart}

\AtBeginDocument{%
  \providecommand\BibTeX{{%
    \normalfont B\kern-0.5em{\scshape i\kern-0.25em b}\kern-0.8em\TeX}}}

\setcopyright{none}

\acmJournal{TOG}
\acmVolume{XX}
\acmNumber{XX}
\acmArticle{XX}
\acmMonth{03}
\acmSubmissionID{XX-XX}

\usepackage{color}
\usepackage{xspace}
\usepackage[normalem]{ulem}
\usepackage[nolist]{acronym}
\usepackage{cleveref}
\usepackage[textfont=footnotesize,labelfont=footnotesize]{subcaption}
\usepackage{bm}
\usepackage{csquotes}
\usepackage[linesnumbered,ruled,vlined]{algorithm2e}
\usepackage{tikz}
\usetikzlibrary{shadows.blur}
\usetikzlibrary{positioning}
\usetikzlibrary{arrows.meta}
\usetikzlibrary{positioning}
\usetikzlibrary{spy}
\usepackage{layouts}
\usepackage{amsmath}
\usepackage{mathtools}
\include{macros}

 \Crefname{section}{Sec.}{Secs.}
 \Crefname{table}{Tab.}{Tabs.}
 \Crefname{figure}{Fig.}{Figs.}
 \Crefname{equation}{Eqn.}{Eqns.}

\begin{document}

\title{Low-light Image and Video Enhancement via Selective Manipulation of Chromaticity}

\author{Sumit Shekhar}
\email{sumit.shekhar-at-hpi.uni-potsdam.de}
\orcid{0000-0002-5683-2290}
\affiliation{%
  \institution{Hasso Plattner Institute, University of Potsdam}
  \country{Germany}
}

\author{Max Reimann}
\orcid{0000-0003-2146-4229}
\affiliation{%
	\institution{Hasso Plattner Institute, University of Potsdam}
	\country{Germany}
}

\author{Amir Semmo}
\orcid{0000-0002-1553-4940}
\affiliation{%
	\institution{Digital Masterpieces GmbH}
	\country{Germany}
}

\author{Sebastian Pasewaldt}
\affiliation{%
	\institution{Digital Masterpieces GmbH}
	\country{Germany}
}

\author{J{\"u}rgen D{\"o}llner}
\affiliation{%
	\institution{Hasso Plattner Institute, University of Potsdam}
	\country{Germany}
}

\author{Matthias Trapp}
\orcid{0000-0003-3861-5759}
\affiliation{%
	\institution{Hasso Plattner Institute, University of Potsdam}
	\country{Germany}
}

\renewcommand{\shortauthors}{Shekhar, et al.}

\begin{abstract}
Image acquisition in low-light conditions suffers from poor quality and significant degradation in visual aesthetics.
This affects the visual perception of the acquired image and the performance of various computer vision and image processing algorithms applied after acquisition. 
Especially for videos, the additional temporal domain makes it more challenging, wherein we need to preserve quality in a temporally coherent manner.
We present a simple yet effective approach for low-light image and video enhancement.
To this end, we introduce "Adaptive Chromaticity", which refers to an adaptive computation of image chromaticity.
The above adaptivity allows us to avoid the costly step of low-light image decomposition into illumination and reflectance, employed by many existing techniques. 
All stages in our method consist of only point-based operations and high-pass or low-pass filtering, thereby ensuring that the amount of temporal incoherence is negligible when applied on a per-frame basis for videos.   
Our results on standard low-light image datasets show the efficacy of our algorithm and its qualitative and quantitative superiority over several state-of-the-art techniques. 
For videos captured in the wild, we perform a user study to demonstrate the preference for our method in comparison to state-of-the-art approaches.  
\end{abstract}



\definecolor{dkgreen}       {rgb}{0.0,0.5,0.0}
\DefAuthor{MT}{dkgreen}

\definecolor{dkred}       {rgb}{0.5,0.0,0.0}
\DefAuthor{AS}{dkred}

\definecolor{dkgelb}       {rgb}{0.3,0.01,0.06}
\DefAuthor{MxR}{dkgelb}

\definecolor{dkblue}       {rgb}{0.0,0.0,0.7}
\DefAuthor{SP}{dkblue}

\definecolor{dkmag}       {rgb}{0.8,0.0,0.8}
\DefAuthor{SSh}{dkmag}

\newcommand\mycommfont[1]{\footnotesize\ttfamily\textcolor{blue}{#1}}
\SetCommentSty{mycommfont}

\SetKwInput{KwInput}{Input}                
\SetKwInput{KwOutput}{Output}              

\maketitle


\input{alive_introduction}

\input{alive_related_work}
\input{alive_method}
\input{alive_results}
\input{alive_discussion}
\input{alive_conclusion}

\begin{acronym}
    \acro{AC}{Adaptive Chromaticity}
	\acro{HE}{Histogram Equalization}
	\acro{DAC}{Denoised Adaptive Chromaticity}
	\acro{HDR}{High Dynamic Range}	
	\acro{NLM}{Non-local Means}
	\acro{LLIE}{Low-light Image Enhancement}
	\acro{LOE}{Lightness Order Error}
	\acro{DSFD}{Dual Shot Face Detector}
	\acro{VES}{Virtual Exposure Sequence}
\end{acronym}

\bibliographystyle{ACM-Reference-Format}
\bibliography{alive}

\end{document}

%% file: macros.tex
\makeatletter
\DeclareRobustCommand\onedot{\futurelet\@let@token\@onedot}
\def\@onedot{\ifx\@let@token.\else.\null\fi\xspace}
\def\etal{\textit{et~al}\onedot~}
\def\eg{e.g\onedot, } 
\def\ie{i.e\onedot, }

\makeatother

\def\clap#1{\hbox to 0pt{\hss #1\hss}}%
\def\initials#1{\protect\clap{\smash{\raisebox{1.4ex}{\tiny{\textsf{\textit{#1}}}}}}}%
\makeatletter
\newcommand{\EDIT}[4][]{\protect\@ifundefined{hidecomments}{%
  \strut{\color{#3}{\hspace{0pt}\initials{#2}\protect\sout{#1}{~#4}}}%
  }{}}
\newcommand{\NOTEboxed}[3]{\protect\@ifundefined{hidecomments}{%
  {\begin{center}\fbox{\parbox{0.97\linewidth}{\protect\EDIT{#1}{#2}{#3}}}\end{center}}
  }{}}
\makeatletter
\newcommand{\DefAuthor}[2] 
{%
  \expandafter\newcommand\csname #1edit\endcsname[2][]{\protect\EDIT[##1]{#1}{#2}{##2}}
  \expandafter\newcommand\csname #1\endcsname[1]{\protect\csname #1edit\endcsname{[##1]}}
  \expandafter\newcommand\csname #1boxed\endcsname[1]{\NOTEboxed{#1}{#2}{##1}}
}
\makeatother



%% file: alive_introduction.tex
\section{Introduction}
\label{RL:Sec:Introduction}

\newcommand{\teaserimagewithspy}[1]{
	\begin{tikzpicture}[every node/.style={inner sep=0,outer sep=0}]
		\begin{scope}[
	   node distance = 11mm,
		   inner sep = 0pt,spy using outlines={rectangle, red, magnification=4 }
		 ]
	   \node (n0)  { \includegraphics[width=\columnwidth]{#1}};
	   \spy [red,size=0.6cm,line width=0.8pt] on (0.18,0.53) in node[anchor=north east,inner sep=0pt] at (n0.north east);
	   \end{scope}
	   \end{tikzpicture}
}
\begin{figure} [tb]
\begin{subfigure}{0.322\linewidth}%
\caption{Input image}%
\teaserimagewithspy{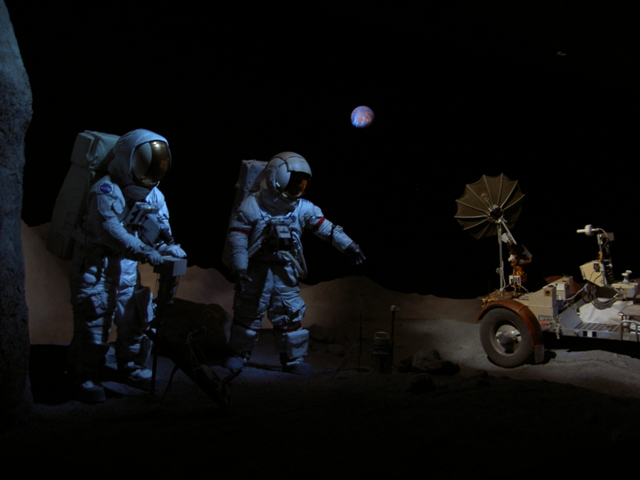}%
\label{fig:teaser_input}%
\end{subfigure}\hfill
\begin{subfigure}{0.322\linewidth}%
\caption{LIME~\shortcite{LIME_Guo2017}}%
\teaserimagewithspy{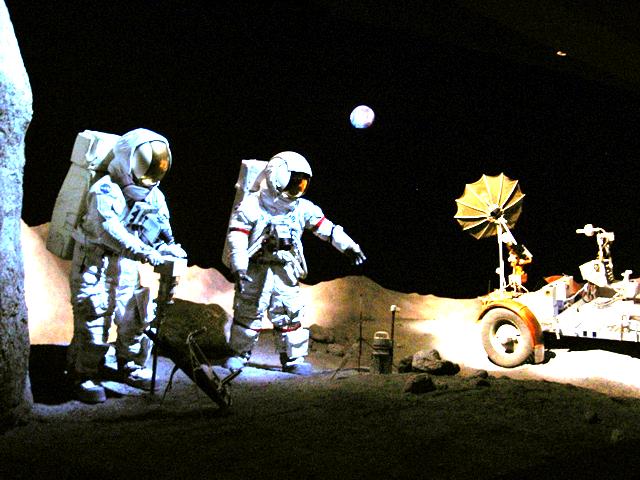}%
\label{fig:teaser_LIME}%
\end{subfigure}\hfill
\begin{subfigure}{0.322\linewidth}%
\caption{MBLLEN~\shortcite{MBLLEN_Lv2018}}%
\teaserimagewithspy{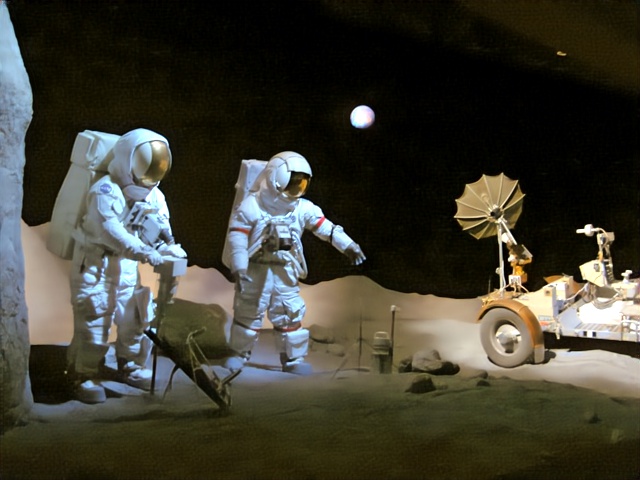}%
\label{fig:teaser_MBLLEN}%
\end{subfigure}\\[-2.7ex]
\begin{subfigure}{0.322\linewidth}%
\caption{Zero-DCE~\shortcite{Zero_DCE_Guo2020}}%
\teaserimagewithspy{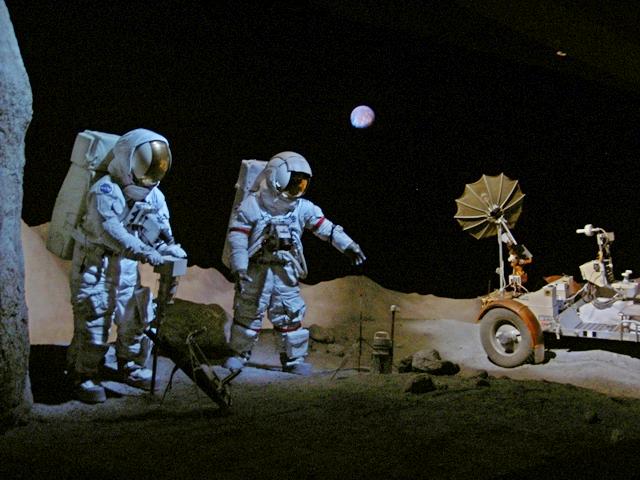}%
\label{fig:teaser_ZeroDCE}%
\end{subfigure}\hfill
\begin{subfigure}{0.322\linewidth}%
\caption{LLVE~\shortcite{Zhang_Temporal2021}}%
\teaserimagewithspy{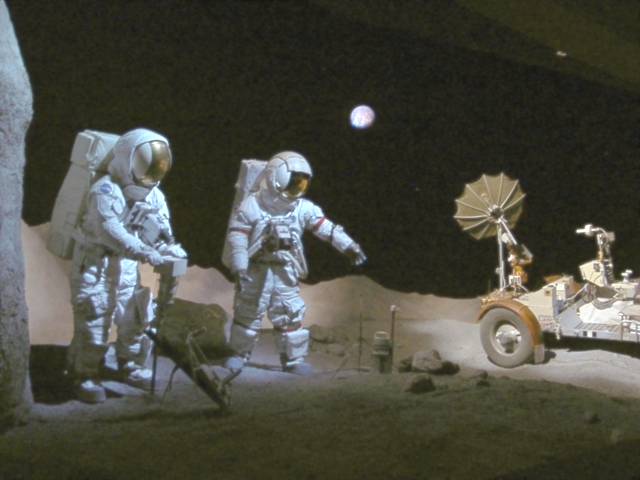}%
\label{fig:teaser_LLVE}%
\end{subfigure}\hfill
\begin{subfigure}{0.322\linewidth}%
\caption{Ours}%
\teaserimagewithspy{resources/teaser/teaser_new/out_ours.png}%
\label{fig:teaser_Ours}%
\end{subfigure}%
\caption{Comparison of \acs{LLIE} results for three image-based (b to d) and one video based (e) method. Our method (f) can brighten image while preserving details and avoiding artifacts in terms of over-exposedness, noise, and desaturation.}%
\label{fig:teaser}%
\end{figure}

Due to unavoidable technical or environmental constraints, images and videos captured in poor lighting conditions suffer from severe degradation of visual information and aesthetic quality.
Furthermore, it is challenging for such visual media to be used for high-level tasks such as object detection or tracking due to a lack of visual information.
Further, poor visual quality negatively affects the visual experience of end-users.

Numerous algorithms have been proposed for \ac{LLIE} (\Cref{fig:teaser}) and a few for video enhancement as well.
A class of methods is based on Retinex theory, which assumes the image to be a product of illumination and reflectance. 
Most of the existing Retinex-based approaches decompose the image into illumination and/or reflectance components, based on specific prior(s).
However, finding an effective prior is challenging and inaccuracies can result in artifacts and color deviations in the enhanced output.
Further, the runtime for such a decomposition, employing a complex optimization process, is relatively long~\cite{Survey_Liu2021}.
In comparison, deep learning-based solutions are faster than conventional methods and learn the underlying prior using the given data distribution.
However, they tend to suffer from limited generalization capability. 
The above could be due to limited/synthetic training data, ineffective network structures, or unrealistic assumptions~\cite{Survey_Li2021}. 
Therefore, we aim to develop a practical solution for \ac{LLIE} which adapts to different low-light conditions and also has low computational complexity for interactive performance on commodity hardware. 
 
To achieve the above objective, we adopt a simple strategy based on Retinex theory, the basis for various conventional and learning-based methods. 
We avoid the computationally costly decomposition step and propose an adaptive way to slowly transition into baseline-reflectance (\ie chromaticity)~\cite{Intrinsic_Bonneel2017}.
We refer to it as \textit{\ac{AC}}, which forms the basis for our approach. 
The adaptive transition into chromaticity can efficiently increase the output brightness while being robust against dark (or low-intensity) pixels.
Further, it prevents amplification of sensor noises, to a large degree, common in low-light images.  
With respect to dark pixels, our approach consistently produces better results for both low and very-low lighting conditions. 
We generate multiple such \acp{AC} with varying level of brightness followed by a multi-scale fusion step.
Different levels of brightness prevents over/under-exposedness while multi-scale fusion preserves fine image details. 

Unlike images, low-light video enhancement has received less attention. 
Application of image-based methods to videos on a per- frame basis is temporally incoherent and often leads to flickering artifacts. 
Dark pixels significantly contribute in noise amplification leading to temporal incoherence. 
Due to our ability to robustly handle such pixels the amount of temporal incoherence is reduced significantly.
Even the per-frame application of our image-based solution is superior to an existing video-specific approach.
Our contributions are summarized as follows, we propose:
\begin{enumerate}
\item \textit{Adaptive Chromaticity} to efficiently increase image brightness while preventing amplification of noise. 
\item An approach for low-light image enhancement based on exposure fusion of various \acp{AC} of the given image.
\item An per-frame application of our image-based solution for videos, which works out-of-the-box without introducing significant temporal incoherence. 
\end{enumerate}

%% file: alive_related_work.tex
\section{Background and Related Work}
\label{RL:Sec:RelatedWork}

\paragraph{Low-Light Enhancement of Images}

One of the earliest algorithms for low-light image enhancement is based on Retinex theory.  
Jobson \etal \shortcite{SSR_Jobson1997,MSR_Jobson1997} propose center/surround Retinex at single-scale and multi-scale to achieve plausible results for dynamic range compression and color restoration.
Various follow-up methods employ Retinex theory as their basis and propose complex optimization strategies to estimate reflectance and/or illumination for the purpose of low-light image enhancement \cite{Naturalness_Wang2013, Probabilistic_Fu2015, Weighted_Fu2016, LIME_Guo2017, Joint_Cai2017, Structure_Li2018, Hybrid_Fu2019, Dual_Zhang2019, LR3M_Ren2020}. 
Fu \etal \shortcite{Weighted_Fu2016} propose a weighted variational model for simultaneous reflectance and illumination
estimation.
Guo \etal \shortcite{LIME_Guo2017} perform refinement of an initial illumination map via a structure prior to obtain a well constructed illumination map thereby enabling enhancement.  
Ren \etal \shortcite{LR3M_Ren2020} propose a robust model to estimate reflectance and illumination maps simultaneously, with provision to suppress noise in the reflectance map. 
Most of the above techniques have long run-time involving CPU-based complex optimization solving for image decomposition. 
We also use the Retinex image formation model as our premise.
However, unlike existing techniques we do not perform the decomposition of image into reflectance and/or illumination layers, thus, achieving interactive performance on commodity hardware. 

Another class of methods for low-light image enhancement is based on \ac{HE}, wherein the histogram of the input image is stretched thereby improving its contrast \cite{AHE_Pizer1987}. 
Similar to Retinex-based approaches, various extension to the basic principle have been proposed \cite{Simple_Cheng2004, Dynamic_Abdullah2007, Contextual_Celik2011, Contrast_Lee2013}.
Celik and Tjahjadi \shortcite{Contextual_Celik2011} employ a variational approach for contrast enhancement using inter-pixel contextual information. 
Lee \etal \shortcite{Contrast_Lee2013} use a layered difference of 2D histograms and thus achieve better results than previous \ac{HE}-based approaches.
However, the primary focus of \ac{HE}-based methods is contrast enhancement instead of physically-based illumination editing, thus having the potential risk of over- and/or under- exposed pixels.

Recently, deep learning has also been used substantially to tackle the problem of low-light image enhancement. 
Methods based on various learning strategies, such as supervised~\cite{LLNet_Lore2017, MBLLEN_Lv2018, Deep_Ret_Wei2018, Learning_ME_Cai2018, Hybrid_Ren2019, KinD_Zhang2019, Learning_Xu2020, EEMEFN_Zhu2020}, semi-supervised~\cite{Fidelity_Yang2020}, unsupervised~\cite{Zero_DCE_Guo2020, EnlightenGAN_Jiang2021, Unsupervised_Lee2020}, and reinforcement learning~\cite{DeepExposure_Yu2018} have been proposed.
Lore \etal~\shortcite{LLNet_Lore2017} present the first deep learning-based method in this context (LLNet) that employs stacked-sparse denoising autoencoder to lighten and denoise low-light images simultaneously.
Lv \etal~\shortcite{MBLLEN_Lv2018} propose an end-to-end multibranch network for simultaneous enhancement and denoising.
Ren \etal~\shortcite{Hybrid_Ren2019} design an encoder-decoder network for global image enhancement and a separate recurrent neural network for further edge enhancement. 
Similar to Ren \etal, Zhu \etal~\shortcite{EEMEFN_Zhu2020} propose a method called EEMEFN, which consists of two
stages: multi-exposure fusion and edge enhancement.
Wang \etal~\cite{Underexposed_Wang2019} propose a network called DeepUPE to model image-to-image illumination and collect an expert-retouched dataset.
Zhang \etal~\cite{KinD_Zhang2019} propose a network called KinD based on Retinex theory and design a restoration module to counterbalance noise. 
Chen \etal~\cite{Learning_Chen2018} collect a dataset named SID and train a U-Net \shortcite{UNet_Ronneberger2015} to estimate enhanced sRGB images from raw low-light images. 
Although learning-based methods can produce visually plausible results, they have limited generalization capability in comparison to conventional methods~\cite{Survey_Li2021}.
Two methods which are closely related to our approach are that of Ying \etal~\shortcite{Bio_Ying2017} and Zheng \etal~\shortcite{Single_Zheng2020}, both generate multiple images with different exposures followed by exposure fusion. 
Ying \etal employ a complex strategy with multiple steps to generate the exposure sequence followed by a computationally expensive optimization solving for fusion.
The exposure sequence generation for Zheng \etal is relatively simpler than above, however, they make use of deep-learning to further enhance the sequence as an intermediate step. 
In comparison, our exposure sequence generation is quite straightforward and does not require any learning-based post-processing. 

Apart from the above, existing techniques when applied on a per-frame basis, \eg for videos, usually suffer from temporal incoherence.
We prevent such inconsistency to a large degree by resorting to only point-based operations and high- or low- pass filtering.

\begin{figure*}[tb]
\centering
\includegraphics[trim={0.1cm 9.8cm 0.6cm 0cm},clip,width=0.935\textwidth]{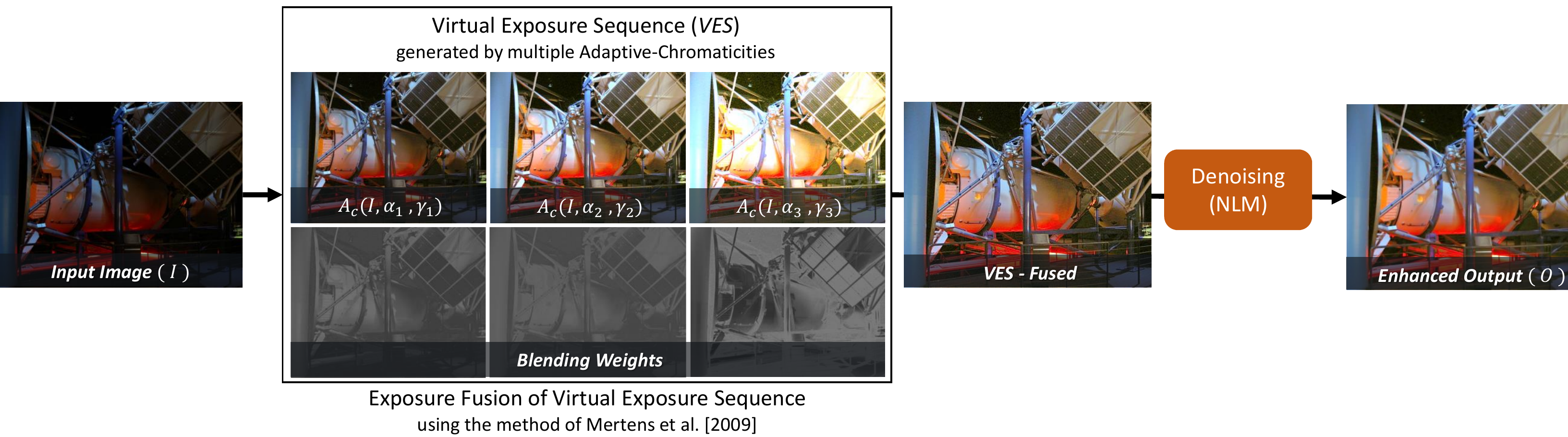}
\caption{Flowchart of our low-light image enhancement algorithm. We generate multiple \acfp{AC} (\Cref{RL:SubSec:AdaptiveChromaticity}) using the input image to create a \acf{VES}~(\Cref{RL:Para:VES_Generation}). As the next step, we blend these images guided by quality measures of contrast, saturation, and well-exposedness~(\Cref{RL:Para:VES_Fusion}). The above is performed in a multi-resolution fashion, as proposed by Mertens \etal~\shortcite{Exposure_Mertens2009}. Finally, we denoise the resulting output to remove remaining noise.}
\label{fig:flow_diagram}
\end{figure*}

\paragraph{Low-Light Enhancement of Videos}

In comparison to images, low-light video enhancement has received significantly less attention. 
One straightforward way to do so would be to stabilize a per-frame based application of low-light image enhancement technique using blind video consistent filtering approaches \cite{Bonneel_Blind2015, Lai_Learning2018, Shekhar_Consistent2019}. 
These techniques inherently make use of vision-based attributes such as optical flow~\cite{Bonneel_Blind2015, Lai_Learning2018} or saliency masks~\cite{Shekhar_Consistent2019} for temporal stabilization.  
However, computation of above vision-based attributes itself will be potentially inaccurate/challenging for low light videos.
Lv \etal~\shortcite{MBLLEN_Lv2018} propose an extension for their learning based approach for images by replacing their 2D convolution layers with 3D ones and train it on synthetic video data.
In order to collect real-world training data, Chen \etal~\shortcite{Chen_SMID2019} capture videos for static scenes with the corresponding long-exposure ground truths and ensure generalization for dynamic scenes by using a Siamese network. 
Jian and Zheng~\shortcite{Jiang_Learning2019} develop a setup to capture bright and dark dynamic video pairs and subsequently train it using a modified 3D U-Net. 
However, their sophisticated setup -- consisting of two cameras, a relay lens and a beam splitter -- is difficult for general usage in the wild.
Triantafyllidou \etal~\shortcite{Triantafyllidou_Low2020} propose a low-light video synthesis pipeline (SIDGAN) that maps \enquote{in the wild} videos into a corresponding low-light domain.
The above approach employs a semi-supervised dual CycleGAN to produce dynamic video data (RAW-to-RGB) with intermediate domain mapping. 
In a recent work, Zhang \etal~\shortcite{Zhang_Temporal2021} enforce temporal stability for low-light video enhancement by predicting optical flow for a single image and synthesizing short range video sequences.
However, their quality of enhancement is low in comparison to existing techniques~(\Cref{RL:SubSec:Quantitative}).
We do not perform any temporal processing specific for videos, however our low-light image enhancement algorithm introduces only negligible temporal incoherence.

%% file: alive_method.tex
\begin{figure} [tb]
	\begin{subfigure}{0.495\columnwidth}%
		\caption{Input image}%
		\includegraphics[trim={0cm 1cm 0cm 0cm},clip,width=\columnwidth]{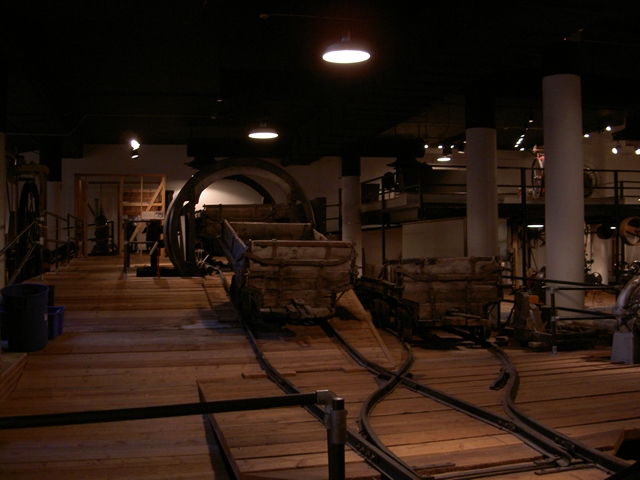}%
		\label{fig:diff_input}%
	\end{subfigure}\hfill
	\begin{subfigure}{0.495\columnwidth}%
		\caption{Chromaticity}%
		\includegraphics[trim={0cm 1cm 0cm 0cm},clip,width=\columnwidth]{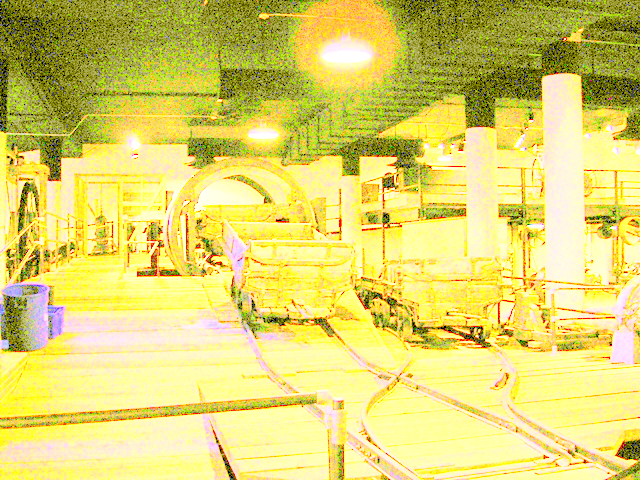}%
		\label{fig:diff_chroma}%
	\end{subfigure}\\[-0.5ex]
	\begin{subfigure}{0.495\columnwidth}%
		\caption{Intensity Difference $y$}%
		\includegraphics[trim={0cm 1cm 0cm 0cm},clip,width=\columnwidth]{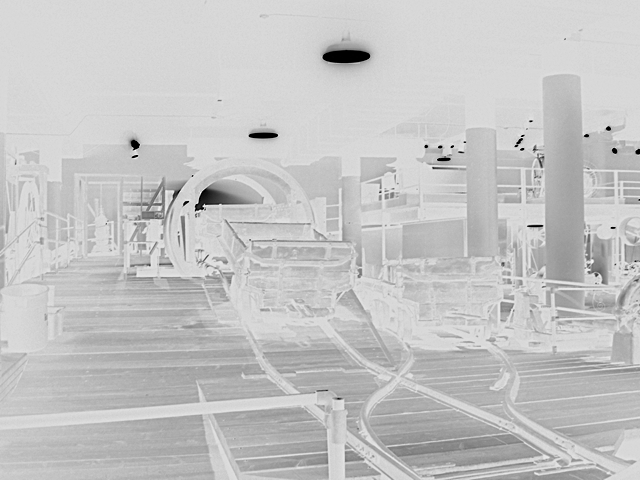}%
		\label{fig:diff_diff}%
	\end{subfigure}\hfill
	\begin{subfigure}{0.495\columnwidth}%
		\caption{Adaptive Chromaticity}%
		\includegraphics[trim={0cm 1cm 0cm 0cm},clip,width=\columnwidth]{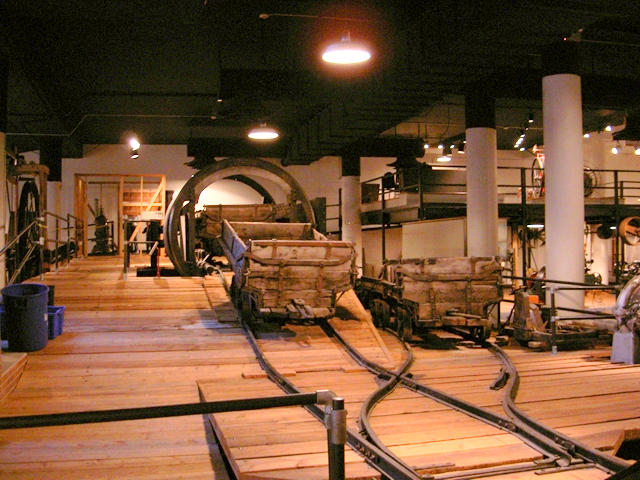}%
		\label{fig:diff_adap_chroma}%
	\end{subfigure}%
	\caption{Given an (a) input image, the noise in the (b) chromaticity is higher for low-intensity pixels with a larger (c) intensity difference, which is significantly reduced for (d) adaptive chromaticity (with $\alpha = 0.3$ and $\gamma = 0.8$).}%
	\label{fig:diff_vis}%
\end{figure}

\section{Method}
\label{RL:Sec:Method}

According to the Retinex model, an image $I$ can be expressed as the product of a \textit{reflectance} layer $R$ and an \textit{illumination} layer $L$~\cite{Retinex_Land1971}: $I = R \times L$,
where the operator $\times$ denotes pixel-wise multiplication.
As a baseline, image \enquote{intensity} and \enquote{chromaticity} can be considered as the illumination and reflectance layer, respectively~\cite{Intrinsic_Bonneel2017}. 
One can employ different approaches to compute image intensity, such as: norm or the maximum of the individual color channels. 
However, it does not yield desirable results for our purpose of perceptually plausible editing (see supplementary material).  
We consider the luminance (Y-channel in YCbCr color space) as our intensity operator $In(\cdot)$ since this satisfies the above objective.
Chromaticity is correspondingly obtained by dividing the image with its intensity (\Cref{RL:Eqn:Chroma}). 
The above division operation is able to significantly reduce shading and shadows in the scene, which only affects the intensity, thus making the chromaticity relatively brighter than the input image. 
Moreover, it also acts as a normalizing factor for pixel color and saturates it further making it appear perceptually bright. 
For an input image $I$ with color channels $r$, $g$, and $b$ in sRGB color space using 8-bit per channel (\ie 24-bit color depth), we define intensity (following ITU-R BT.601) by the operator $In(\cdot)$ and chromaticity $C$ as follows:
\begin{equation}\label{RL:Eqn:Chroma}
In(I) = 0.299 \cdot r + 0.587 \cdot g + 0.144 \cdot b \; \mathrm{and} \;  C = \frac{I}{In(I)}.
\end{equation}  
The brightening effect of chromaticity is a preferable characteristic for low-light image enhancement.
However, chromaticity suffers from un-desirable artifacts in terms of \textit{noise} and \textit{color-shifts} especially for low-intensity pixels (\Cref{fig:diff_chroma}).

\subsection{Adaptive Chromaticity}
\label{RL:SubSec:AdaptiveChromaticity}

In order to preserve the brightening effect of chromaticity while avoiding artifacts, we introduce \emph{\acf{AC}}.
For identifying a low-intensity pixel, we compute the difference between pixel intensity, $\mathit{In}(\cdot)$, and the maximum intensity value $\mathit{MaxIn}$.
For low-intensity pixels, this difference defined as $y = \mathit{MaxIn} - \mathit{In}(\cdot)$ would be comparatively larger. 
For example, for an intensity image encoded in the range of $0$ to $1$, $MaxIn = 1$ and for a low-intensity pixel $\bm{p}$ with $In(\cdot) = 0.05$ the difference $y(\bm{p}) = 0.95$ is large.
Similarly, for a high-intensity pixel $\bm{q}$ with $\mathit{In}(\cdot) = 0.8$ the difference $y(\bm{q}) = 0.2$ is small~(\Cref{fig:diff_vis}). 
The above forms the basis for defining adaptive chromaticity ($A_c$), wherein we add an adaptive term in the denominator while computing chromaticity (\Cref{RL:Eqn:Chroma}). 
To further increase the brightness, we perform a non-linear scaling using \emph{gamma correction}
\begin{equation}\label{RL:Eqn:AdaptiveChroma}
A_c(I, \alpha, \gamma) = {\left(\frac{I}{\mathit{In}(I) + \alpha (f(y) + h)}\right)}^\gamma.
\end{equation}   
Here, $f(y)$ is a function in terms of $y$, $\alpha$ is a control parameter, $h$ is a small constant, and $\gamma$ is a parameter for gamma correction.
The adaptive function $f(y)$ should be chosen such that its value is close to zero when $y$ is small and is substantially high for significantly large value of $y$.
Thus, by tuning the control parameter $\alpha$ we can smoothly translate between the bright chromaticity (when $\alpha \rightarrow 0$) and a complete dark image (when $\alpha \rightarrow \infty$).
The intuition behind the adaptive denominator in \Cref{RL:Eqn:AdaptiveChroma} is that we divide by a larger value for low-intensity pixels as compared to high-intensity pixels, thereby, reducing undesirable artifacts.
For adaptivity, we can choose a function $f$ which satisfies the above property, we use $f(y) = y^2$ which is efficient to compute and gives plausible results.
The \ac{AC} brightens up an image while significantly reducing these artifacts (\Cref{fig:diff_adap_chroma}) and forms the basis for our low-light image and video enhancement approach.

\begin{figure} [tb]
\begin{subfigure}{0.24\linewidth}%
	\caption{Input image}%
	\includegraphics[trim={0.5cm 0.2cm 0cm 12.0cm},clip,width=\columnwidth]{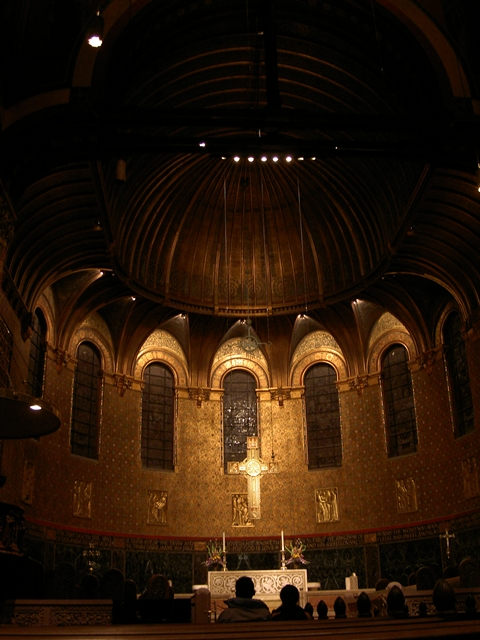}%
	\label{fig:dac_input}%
\end{subfigure}\hfill
\begin{subfigure}{0.24\linewidth}%
	\caption{\ac{DAC}}%
	\includegraphics[trim={0.5cm 0.2cm 0cm 12.0cm},clip,width=\columnwidth]{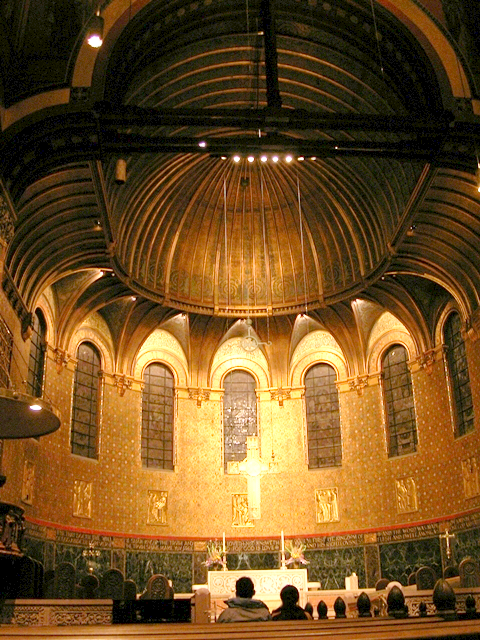}%
	\label{fig:out_dac}%
\end{subfigure}\hfill
\begin{subfigure}{0.24\linewidth}%
	\caption{LIME~\shortcite{LIME_Guo2017}}%
	\includegraphics[trim={0.5cm 0.2cm 0cm 12.0cm},clip,width=\columnwidth]{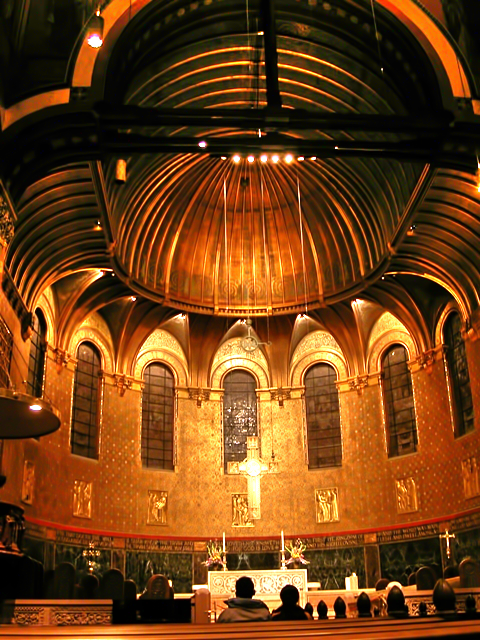}%
	\label{fig:out_lime}%
\end{subfigure}\hfill
\begin{subfigure}{0.24\linewidth}%
	\caption{MBLLEN~\shortcite{MBLLEN_Lv2018}}%
	\includegraphics[trim={0.5cm 0.2cm 0cm 12.0cm},clip,width=\columnwidth]{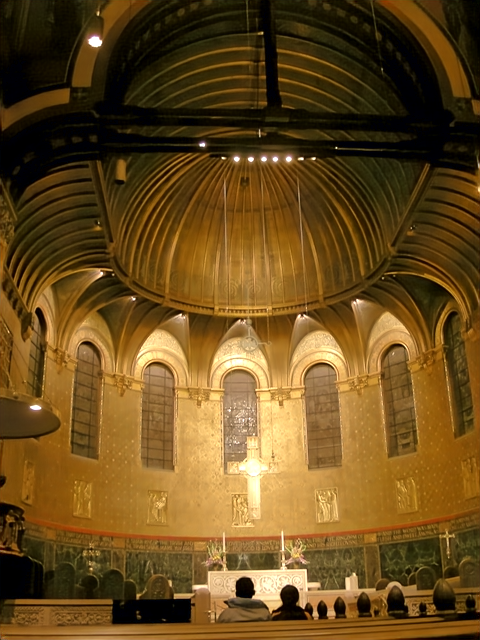}%
	\label{fig:out_mbblen}%
\end{subfigure}%
\caption{The \ac{DAC} is already comparable to state-of-the-art results.}%
\label{fig:dac_results}%
\end{figure}

\begin{figure*}[tb]
	\begin{subfigure}{0.195\linewidth}%
		\caption{Input image}%
		\includegraphics[trim={0.4cm 0.5cm 1cm 0.3cm},clip,width=\linewidth]{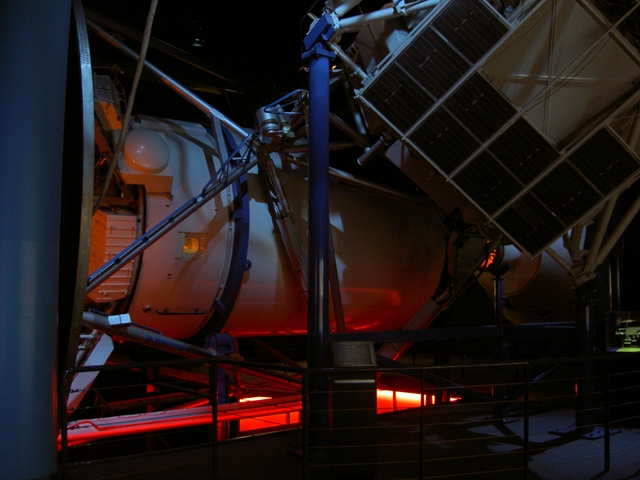}%
		\label{fig:input}%
	\end{subfigure}\hfill
	\begin{subfigure}{0.195\linewidth}%
		\caption{Chromaticity}%
		\includegraphics[trim={0.4cm 0.5cm 1cm 0.3cm},clip,width=\linewidth]{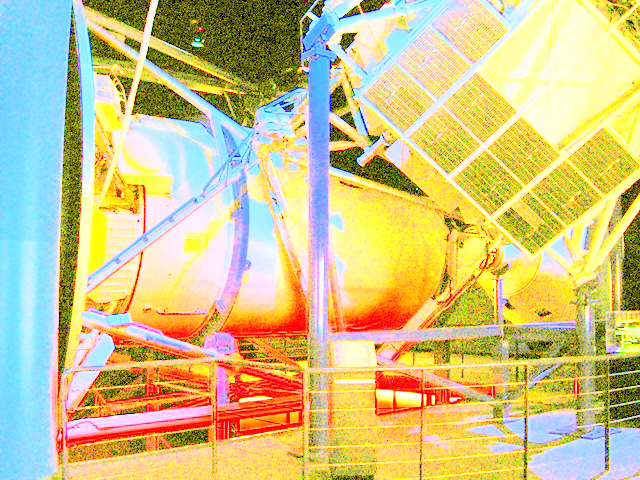}%
		\label{fig:chromaticity}%
	\end{subfigure}\hfill
	\begin{subfigure}{0.195\linewidth}%
		\caption{\ac{AC}}%
		\includegraphics[trim={0.4cm 0.5cm 1cm 0.3cm},clip,width=\linewidth]{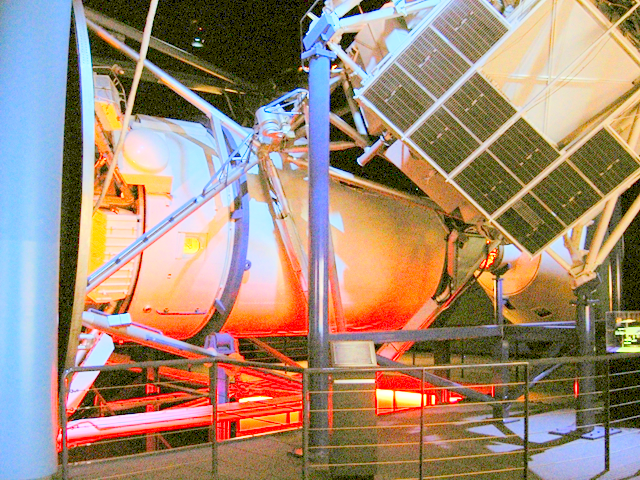}%
		\label{fig:adap_chromaticity}%
	\end{subfigure}\hfill
	\begin{subfigure}{0.195\linewidth}%
		\caption{\ac{DAC}}
		\includegraphics[trim={0.4cm 0.5cm 1cm 0.3cm},clip,width=\linewidth]{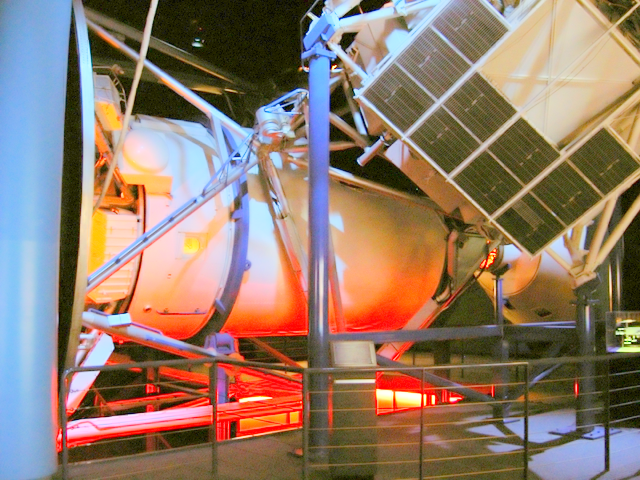}%
		\label{fig:den_adap_chromaticity}%
	\end{subfigure}\hfill
	\begin{subfigure}{0.195\linewidth}%
		\caption{Our Result}%
		\includegraphics[trim={0.4cm 0.5cm 1cm 0.3cm},clip,width=\linewidth]{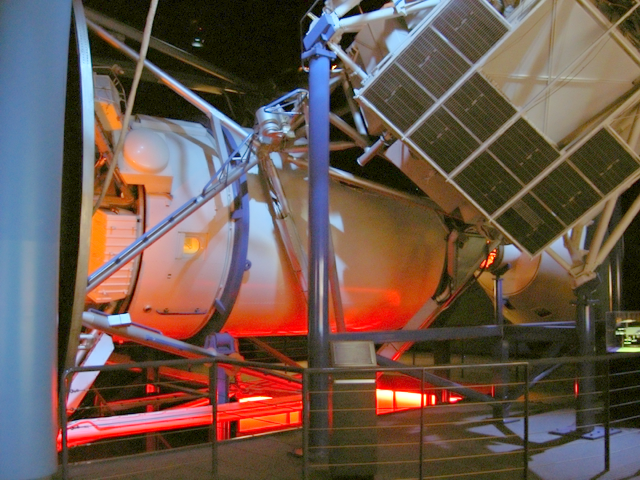}%
		\label{fig:our_result}%
	\end{subfigure}%
	\caption{For (a), low-light image the corresponding (b) chromaticity has artifacts in terms of color-shifts and noise. These artifacts are significantly reduced for (c) \ac{AC} ($\alpha = 0.15, \gamma = 0.6$), the noise can be further removed by a (d) denoising operation. We employ a multi-exposure technique ($3$ exposure levels ($\alpha_1 = 0.85, \alpha_2 = 0.6, \alpha_3 = 0.15$ and $\gamma_1 = \gamma_2 = \gamma_3 = 0.6$) and $4$ pyramid levels) to preserve details (e) by preventing over-exposedness in (d).}%
	\label{fig:adap_chroma}
\end{figure*}

However, some artifacts in terms of sensor noise might still remain, which can be removed by a denoising operation~(\Cref{fig:den_adap_chromaticity}). 
We employ a fast yet effective denoising scheme in the form of \ac{NLM} image denoising~\cite{NLM_Buades2005} for this purpose.  
In principle, one can use more effective denoising techniques such as BM3D~\cite{BM3D_Dabov2007} or FFDNet~\cite{FFDNet_Zhang2018}.
However, the above are not able to handle high-resolution images properly: BM3D is slow in performance and FFDNet requires high-end GPUs.
In comparison, \ac{NLM} is able to provide plausible denoised results for images with varying resolution in reasonable time. 
Thus, the \ac{DAC} obtained is efficient to compute, runs at real-time framerates for high-resolution images ($1920 \times 1080$ pixels), and is already comparable to state-of-the-art methods in terms of quality (\Cref{fig:dac_results}).  
However, since the brightening effect is achieved only by point-based operations without considering the neighborhood, it is unable to preserve fine details especially for bright-saturated regions.

\begin{figure} [tb]
	\begin{subfigure}{0.327\columnwidth}%
		\subcaption{$\gamma = 1.0$ and $\alpha = 0.1$}%
		\includegraphics[width=\columnwidth]{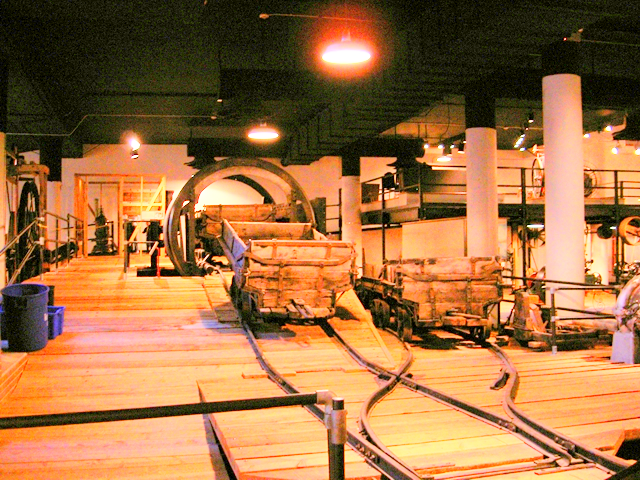}%
		\label{fig:g_1_0_a_0_01}%
	\end{subfigure}\hfill
	\begin{subfigure}{0.327\columnwidth}%
		\subcaption{$\gamma = 1.0$ and $\alpha = 0.5$}%
		\includegraphics[width=\columnwidth]{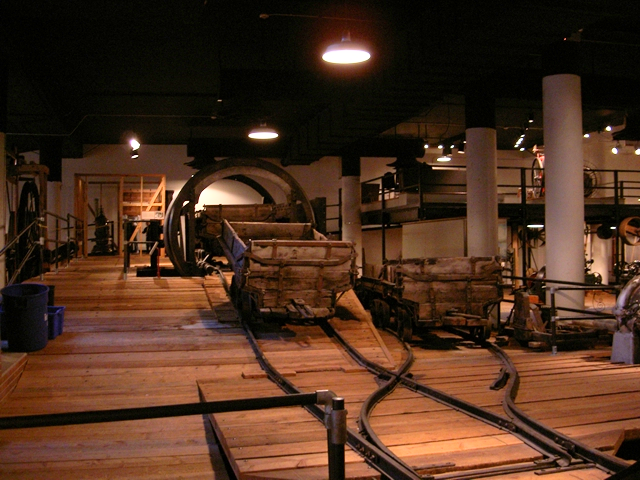}%
		\label{fig:g_1_0_a_0_05}%
	\end{subfigure}\hfill
	\begin{subfigure}{0.327\columnwidth}%
		\subcaption{$\gamma = 1.0$ and $\alpha = 0.9$}%
		\includegraphics[width=\columnwidth]{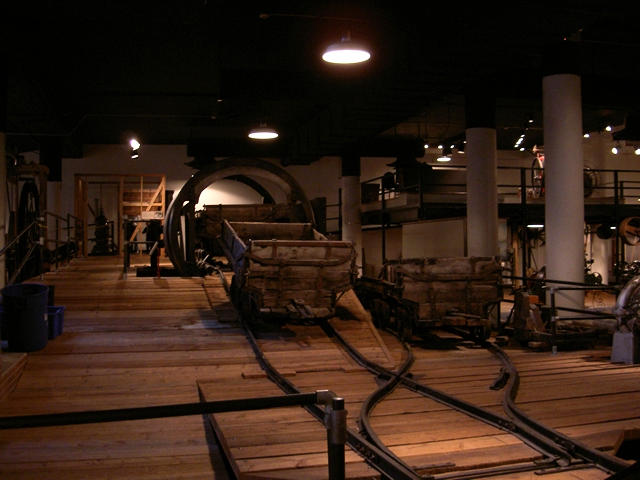}%
		\label{fig:g_1_0_a_0_15}%
	\end{subfigure}\\[-1.5ex]
	\begin{subfigure}{0.327\columnwidth}%
		\subcaption{$\gamma = 0.5$ and $\alpha = 0.1$}%
		\includegraphics[width=\columnwidth]{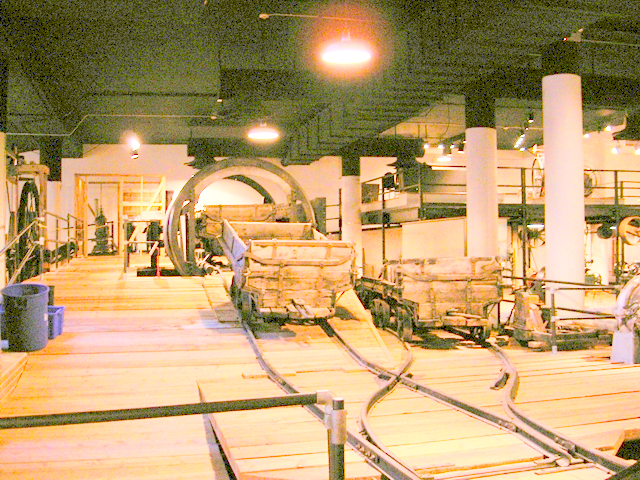}%
		\label{fig:g_0_25_a_0_01}%
	\end{subfigure}\hfill
	\begin{subfigure}{0.327\columnwidth}%
		\subcaption{$\gamma = 0.5$ and $\alpha = 0.5$}%
		\includegraphics[width=\columnwidth]{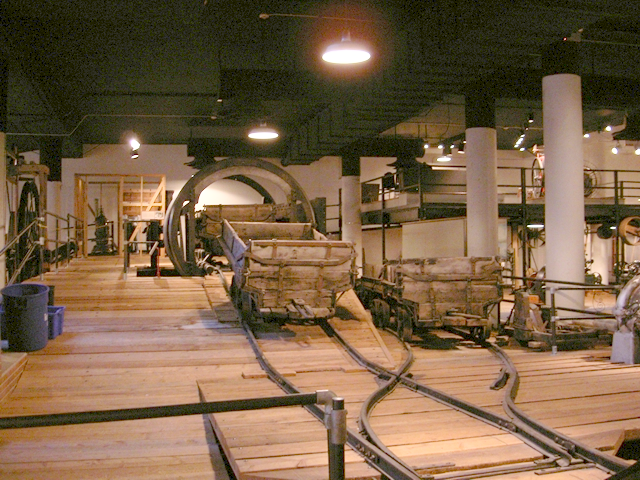}%
		\label{fig:g_0_25_a_0_05}%
	\end{subfigure}\hfill
	\begin{subfigure}{0.327\columnwidth}%
		\subcaption{$\gamma = 0.5$ and $\alpha = 0.9$}%
		\includegraphics[width=\columnwidth]{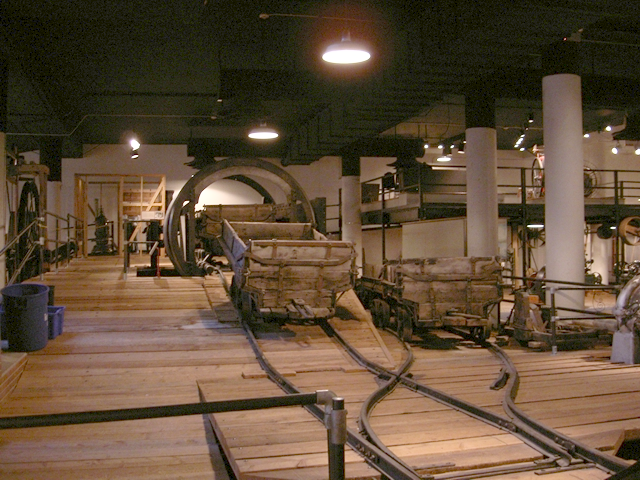}%
		\label{fig:g_0_25_a_0_15}%
	\end{subfigure}\\[-1.5ex]
	\caption{\acf{VES} for the input image in \Cref{fig:diff_vis}: as a sequence of \acp{AC} generated by varying values of $\alpha$ and $\gamma$.}%
	\label{fig:exposure_sequence}
\end{figure}

\subsection{Our Approach for \ac{LLIE}}
\label{RL:SubSec:Our_Approach}

We propose an enhancement approach to overcome the above shortcomings as a two-step process consisting of \acf{VES} \textit{generation} and \textit{fusion}. 
A flowchart of our complete pipeline is depicted in \Cref{fig:flow_diagram}.

\paragraph{\ac{VES} Generation}
\label{RL:Para:VES_Generation}
The overall exposedness of the image is increased by lowering $\alpha$ and/or $\gamma$ values. 
However, the brightening effect of either of these parameters $\alpha$ or $\gamma$ is slightly different. 
For lower values of $\alpha$, increase in brightness comes at the cost of color-shifts~(\Cref{fig:g_1_0_a_0_01}, \Cref{fig:g_0_25_a_0_01}).
On the other hand for lower $\gamma$ values, an increase in brightness is accompanied with desaturation~(\Crefrange{fig:g_0_25_a_0_01}{fig:g_0_25_a_0_15}).
For both $\alpha$ and $\gamma$, lower values leads to increase in noise~(\Cref{fig:g_0_25_a_0_01}) (see supplementary material).
Increasing the exposedness by tuning either $\alpha$ or $\gamma$ is a point-based operation and does not respect the relative contrast within the image. 
The above leads to the problem, wherein already visible regions in the low-light image gets over-exposed while increasing the brightness.
It is similar to challenges in \ac{HDR} photography, which aims to preserve all the details within a \ac{HDR} scene.

We do not have an \ac{HDR} version of the image at our disposal, however we can generate an exposure sequence, with varying values of $\alpha$ and $\gamma$.
One can generate an \ac{HDR} image using the above sequence of images and further tone-map it to preserve details in both bright and dark regions while enhancing it~\cite{HDR_Reinhard2010}. 
Thus, we generate a \emph{virtual exposure sequence} for the given input image by computing \acp{AC} with varying brightness by setting the parameters $\alpha$ and $\gamma$. 
For an image $I$, an exposure sequence $\{E_k|\;k=1 \dots N\}$ is obtained based on the parameter series $\{(\alpha_k, \gamma_k)\,|\,k=1 \dots N\}$, with
\begin{equation}
E_k = A_c(I,\alpha_k, \gamma_k).
\end{equation}

\paragraph{\ac{VES} Fusion}
\label{RL:Para:VES_Fusion}
For efficiency, we skip the step of computing an \ac{HDR} image, and directly fuse the multiple exposures into a high-quality, low dynamic range image using the exposure-fusion technique of Mertens \etal~\shortcite{Exposure_Mertens2009}.
The well-exposedness of an image in the exposure sequence is determined based on quality measures of \textit{contrast} ($c_k$), \textit{saturation} ($s_k$), and \textit{well-exposedness} ($e_k$) on a per-pixel ($\bm{x}$) basis (see supplementary material).
The three quality measures are combined into a joint weighting function
\begin{equation}
	w_k(\bm{x}) = {c_k}^{\upsilon_c}(\bm{x}) \cdot {s_k}^{\upsilon_s}(\bm{x}) \cdot {e_k}^{\upsilon_e}(\bm{x}),
\end{equation}
where the above product can be seen as logical conjunction and the parameters $\upsilon_c$, $\upsilon_s$, and $\upsilon_e$ control the influence of individual quality measures. 
Finally, the obtained sequence of weight maps are normalized such that they sum up to one at each pixel location $\bm{x}$, thereby ensuring consistent results, as follows: 
\begin{equation}
\widehat{w_k}(\bm{x}) = \frac{w_k(\bm{x})}{\sum_{k=1}^{N}w_k(\bm{x})}.
\end{equation}
Once the weight maps are computed, a Laplacian pyramid $\bm{L}(E_k)$ of each input image and a Gaussian pyramid of each normalized weight map $\bm{G}(\widehat{w_k})$ are generated. 
At each pyramid level $l$, the images are fused at per-pixel and per-color channel basis as
\begin{equation}
 {\bm{L}(O)}_l = \sum_{k=1}^{N}{\bm{G}(\widehat{w_k})}_l{\bm{L}(E_k)}_l.
\end{equation}
The final output is obtained by collapsing the computed Laplacian pyramid $\bm{L}(O)$.
Following the above, we employ a denoising operation (similar to \ac{DAC}) to remove any remaining noise. 
All the steps in our method are efficiently summarized in an algorithm in the supplementary material.

%% file: alive_results.tex
\section{Results}
\label{RL:Sec:Results}

\subsection{Parameter Settings} 
\label{RL:SubSec:ParameterSettings}
Our method has three major steps, for which the parameter settings are discussed in the following. 

\paragraph{\ac{VES} Generation}

Ideally, to capture fine details at different exposure levels, multiple images are required for  the exposure sequence. 
However, with increase in number of images processing time will increase accordingly. 
Empirically, we determine three exposure levels ($N = 3$) as sufficient to obtain visually plausible results. 
For any given scene we keep $\gamma$ as constant, thus $\gamma_1 = \gamma_2 = \gamma_3 = \gamma$. 
Empirically, we determine $\gamma \in [0.6, 1.0]$ to give well-exposed and less-noisy results.
For most of our results, we set $\gamma = 0.6$ (for low-noise images) or $\gamma = 0.9$ (for high-noise images).
Empirically, we determine $\alpha \in [0.1, 3.5]$ to yield plausible output.
Unlike $\gamma$, we set three different values of $\alpha$ for a given scene to obtain three different exposure levels respectively.
For most of the results in the paper, we set these as $\alpha_1 = 0.15$ (high-level of brightness), $\alpha_2 = 0.6$ (mid-level of brightness), and $\alpha_3 = 0.85$ (low-level of brightness).
Otherwise we mention the used parameters in the caption or in the supplementary. 

\paragraph{\ac{VES} Fusion and Denoising}

For exposure fusion, we set the weighting exponents for the quality measures to $\upsilon_c = \upsilon_s = \upsilon_e = 1$, as suggested by Mertens~\etal~\shortcite{Exposure_Mertens2009}. 
During fusion, higher number of pyramid-levels helps in preserving fine details.
However, with increase in number of levels processing time increases accordingly which is more pronounced for high-resolution images. 
Empirically, we determine four pyramid levels ($M = 4$) as sufficient to obtain visually plausible results.

For denoising, the \ac{NLM} approach requires two parameters \textit{threshold} ($\mathit{th}$) and \textit{level} ($\mathit{lv}$). 
For us, $\mathit{th} = 0.7$ and $\mathit{lv} = 1.5$ works best for most of the cases. 
Otherwise we mention the used parameters in the caption or in the supplementary. 
On lowering the threshold value significantly, severe denoising leads to loss in details. 

\subsection{Qualitative and Quantitative Evaluation}
\label{RL:SubSec:Qualitative}

We compare our results with state-of-the-art image-based methods: two conventional methods (SRIE~\shortcite{Structure_Li2018} and LIME~\shortcite{LIME_Guo2017}), two supervised-learning based methods (MBBLEN~\shortcite{MBLLEN_Lv2018} and RetinexNet \shortcite{Deep_Ret_Wei2018}), a unsupervised-learning based method (Zero-DCE~\shortcite{Zero_DCE_Guo2020}), and a video-based method (LLVE~\shortcite{Zhang_Temporal2021}). 
The results are produced from publicly available source codes with given parameter settings. 

\paragraph{Images} 

We test the above methods on images taken from the following datasets LIME~\shortcite{LIME_Guo2017}, DICM~\shortcite{Contrast_Lee2013}, NPE~\shortcite{Naturalness_Wang2013}, VV~\shortcite{VV_2022}. 
For quantitative evaluation, we employ the \ac{LOE} metric to compare the performance of different methods on the above datasets. 
\Cref{RL:Tab:eval_loe} shows that we perform better than compared approaches except for MBLLEN. 
However, visually we are able to better preserve the details in comparison to MBLLEN. We provide such comparison for enhanced image outputs in \Cref{fig:qual_result_images}. 
The results of LIME(\Cref{fig:qual_result_images}(b)) tends to be over-exposed, MBLLEN provides satisfactory brightening (\Cref{fig:qual_result_images}(d)) however tends to over-smooth image details, the output of RetinexNet (\Cref{fig:qual_result_images}(e)) do not look natural, and for LLVE the results (\Cref{fig:qual_result_images}(g)) appear to be hazy and desaturated. 
Our results look visually comparable to Zero-DCE and SRIE, however we are able to better preserve details (\eg clouds in the sky in Row-1) and brighten image details in a large dynamic range scenario (\eg human faces in Row-2).     

\begin{table}[t]
\caption{\ac{LOE}~\shortcite{Naturalness_Wang2013} values for images in LIME~\shortcite{LIME_Guo2017}, DICM~\shortcite{Contrast_Lee2013}, NPE~\shortcite{Naturalness_Wang2013}, and VV datasets. The best value is shown in \textcolor{red}{red} and the next best in \textcolor{blue}{blue}.}
\label{RL:Tab:eval_loe}
\begin{tabular}{|l|l|l|l|l|l|}
	\hline
	\small{Method \textbackslash Dataset}     & DICM                                  & LIME                                   & NPE                                & VV & Avg.  \\ \hline
	LIME       & 811.48                                & 709.56                                  & 792.12                                 & 628.67   & 735.46  \\ \hline
	SRIE       & 696.67                             & 681.62                                 & 738.75                                 & 481.57   & 649.65 \\ \hline
	MBLLEN     & \textcolor{red} {\textbf{537.13}}      & \textcolor{red}{\textbf{608.52}}                                 & \textcolor{red}{\textbf{496.99}}                                & \textcolor{red}{\textbf{310.34}}    & \textcolor{red}{\textbf{488.25}}  \\ \hline
	RetinexNet & 836.65                            & 590.10        & 804.09                                 & 774.83   & 751.42  \\ \hline
	LLVE       &  805.24                              & 687.34                                 & 805.59      & 642.01    & 890.75  \\ \hline
	Zero-DCE   & 783.06                                & 697.21                           &796.06                              & 591.39    & 716.93  \\ \hline
	Ours       & \textcolor{blue} {\textbf{638.43}}  & \textcolor{blue}{\textbf{641.12}}   & \textcolor{blue}{\textbf{694.10}} & \textcolor{blue}{\textbf{435.23}}   & \textcolor{blue}{\textbf{602.22}}  \\ \hline
\end{tabular}
\end{table}

\paragraph{Videos}

\begin{figure}[t]
\centering
\includegraphics[trim={5.5cm 2.4cm 6.0cm 4cm},clip,width=1.0\columnwidth]{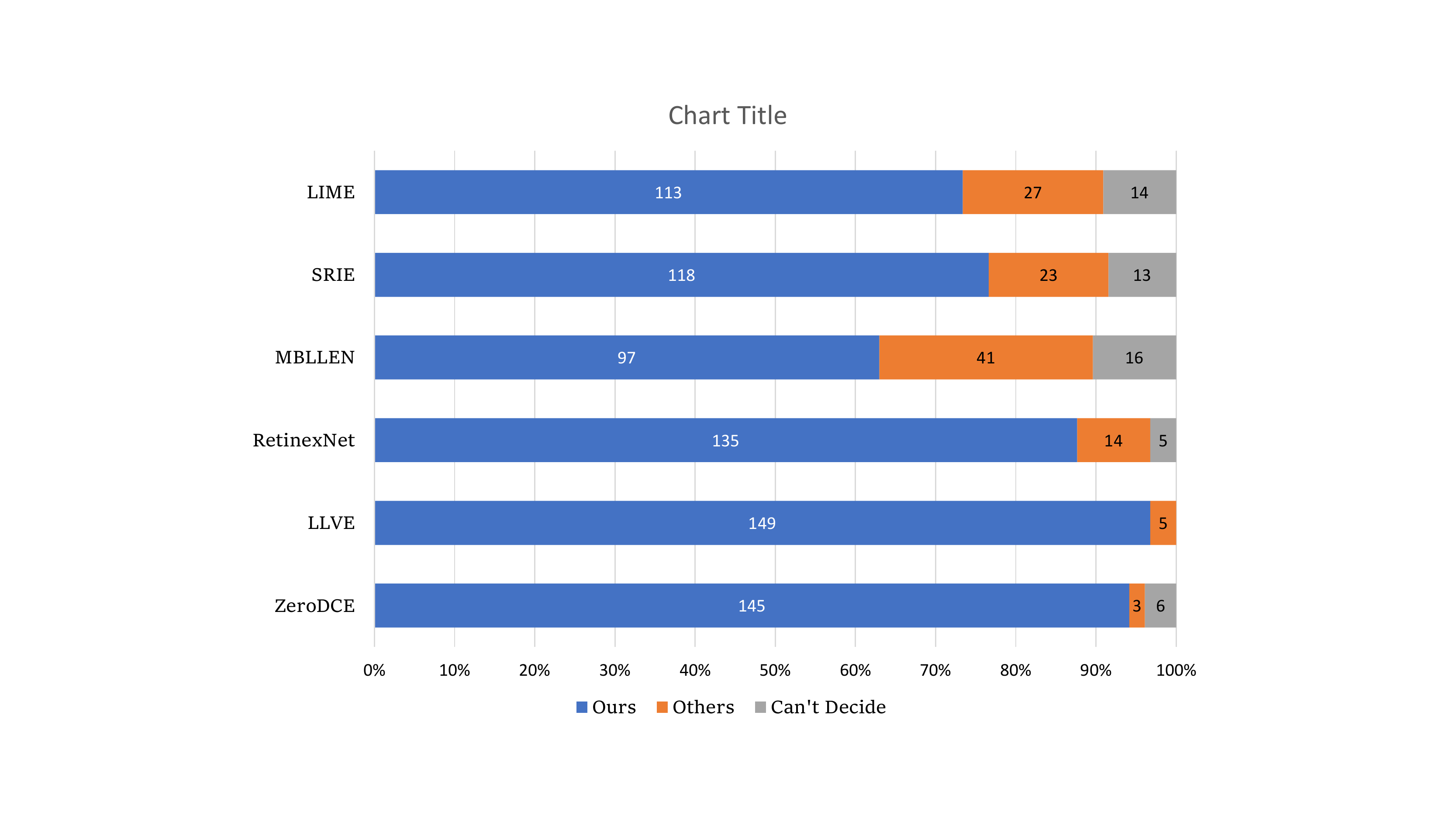}
\caption{Statistics of user study results on low-light video enhancement.}
\label{fig:user_study}
\end{figure}

To evaluate video-enhancement results, we make use of the challenging low-light videos provided by Li~\etal in their survey LLIV~\shortcite{Survey_Li2021}.  
We perform a subjective user study with participants to evaluate the performance of different techniques.
In total, $22$ people ($3$ female, $18$ male, and $1$ non-binary) within the ages of $10$ to $50$ years participated in the study.
The experiment consists of $7$ different low-light videos enhanced by ours and $6$ other ($5$ image-based and $1$ video-based) approaches. 
Two enhanced videos are shown to a participant simultaneously (one of them is ours), thereby constituting $42$ blind A/B tests. 
We asked the participants to focus on the following aspects during comparison:
\begin{description}
	\item[Exposure:] As compared to the input, the output video should be well-exposed, neither under- nor over-exposed.
	\item[Noise and flickering:] The output video should have less noise and flickering. However, the denoising should not be excessive as to remove details. 
	\item[Color:] The color in the output video should appear natural and it should not look over- or under-saturated. 
\end{description}
\Cref{fig:user_study} shows that our method surpasses all other methods including LLVE by a large margin. 

\begin{figure}[tb]
\centering 
\includegraphics[width=0.98\linewidth]{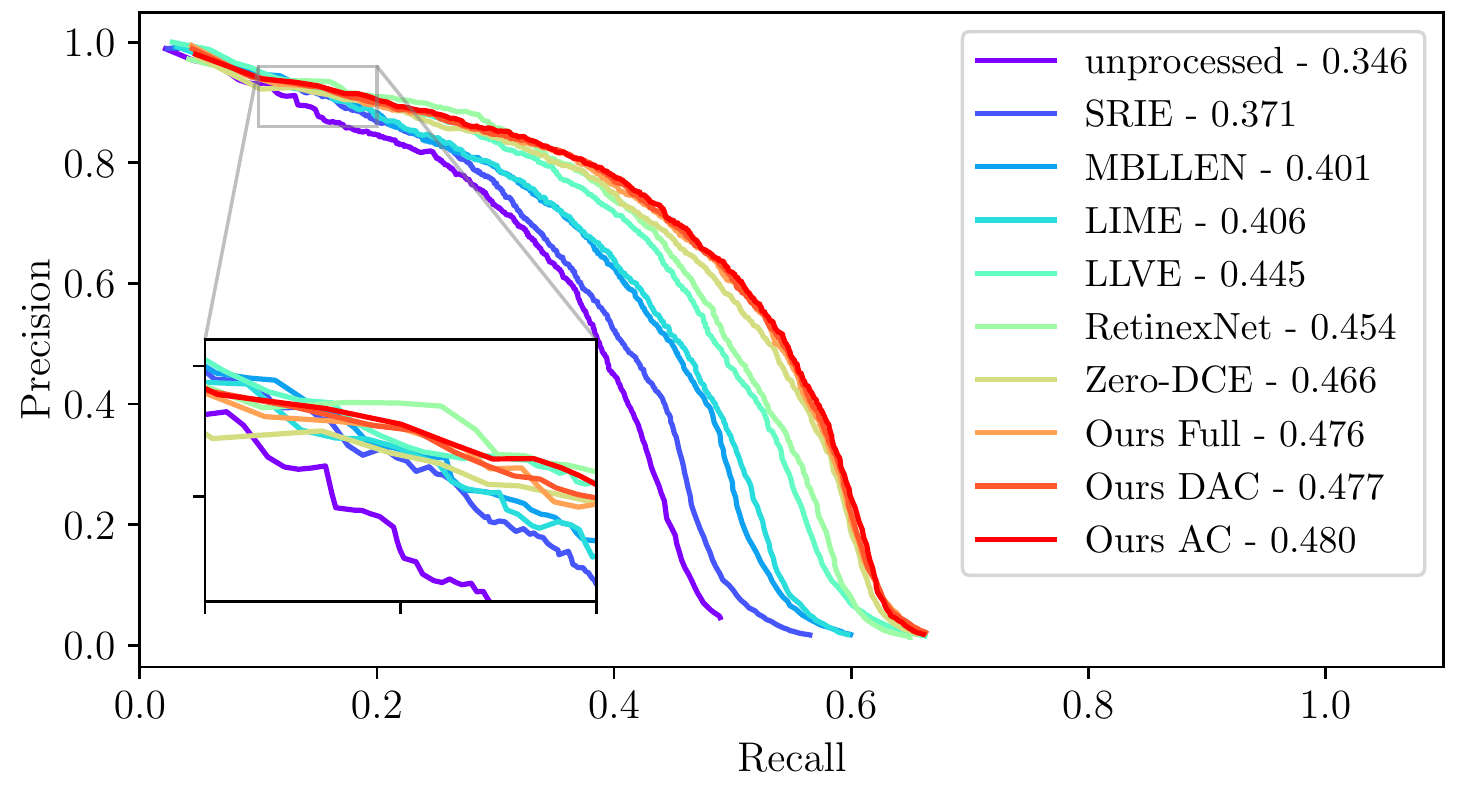}
\caption{Precision-recall curves for face detection on dark-face images \shortcite{poor_visibility_benchmark} enhanced using different \ac{LLIE} methods. Average precision (AP) of each method is indicated in the legend. All our variants set $\alpha=0.25, \gamma=0.6$.}
\label{fig:facedetection_PR_curve}
\end{figure}

\newcommand{\imagewithspy}[3]{
    \begin{tikzpicture}[every node/.style={inner sep=0,outer sep=0}]
        \begin{scope}[ 
			inner sep = 0pt,outer sep=0,spy using outlines={rectangle, red, magnification=2}
                ]
        \node (n0)  { \includegraphics[width=\linewidth]{#1}};
        \spy [red,size=0.6cm] on (#2,#3) in node[anchor=south west,inner sep=0pt] at (n0.south west);
        \end{scope}
    \end{tikzpicture}
}
\begin{figure*}[tb]
	\begin{subfigure}{0.122\linewidth}%
		\subcaption{\footnotesize{Input}}%
		\imagewithspy{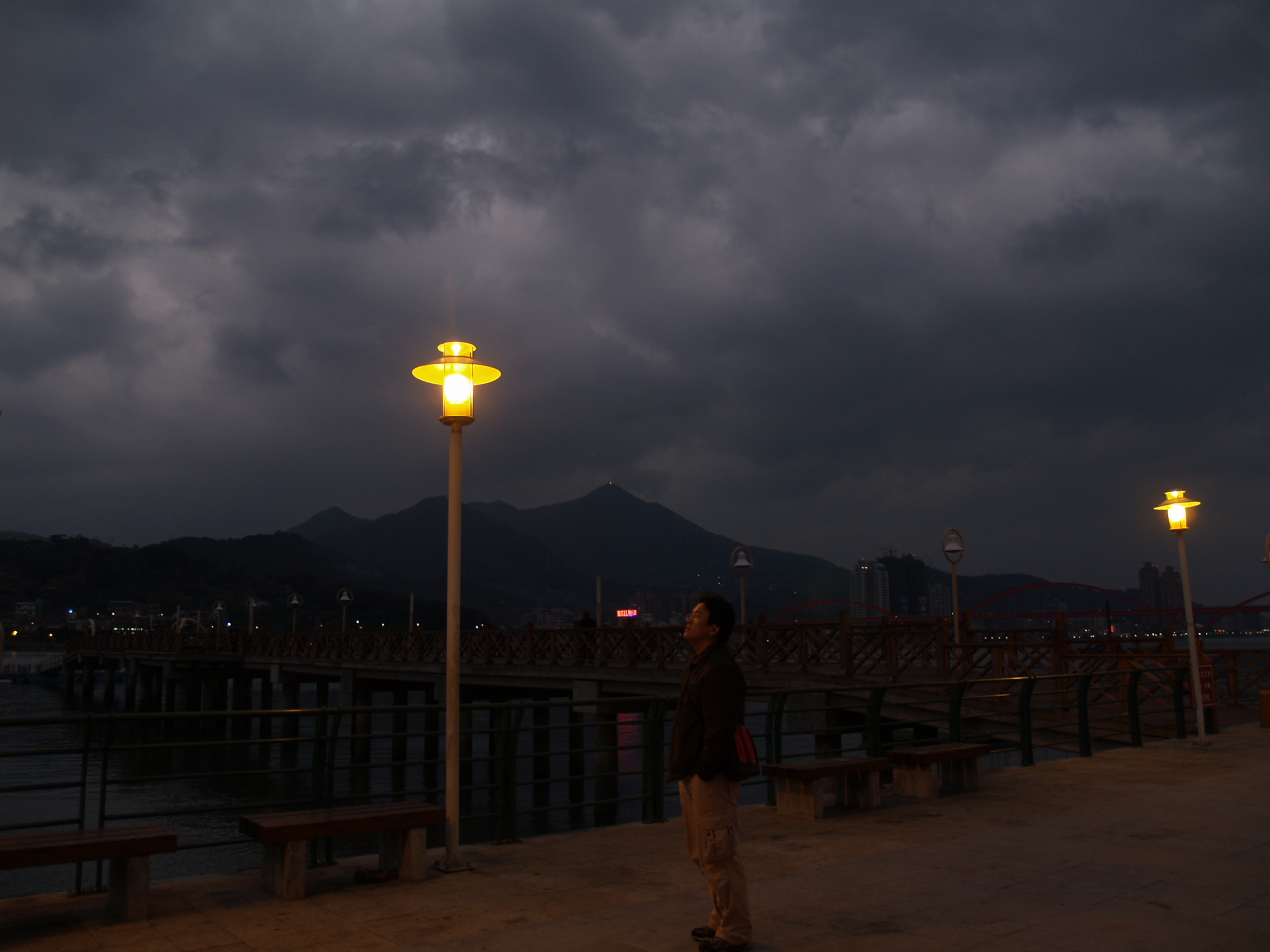}{-0.85cm}{0.4cm}%
		\label{fig:input_1}%
	\end{subfigure}\hfill
	\begin{subfigure}{0.122\linewidth}%
		\subcaption{\footnotesize{LIME~\shortcite{LIME_Guo2017}}}%
		\imagewithspy{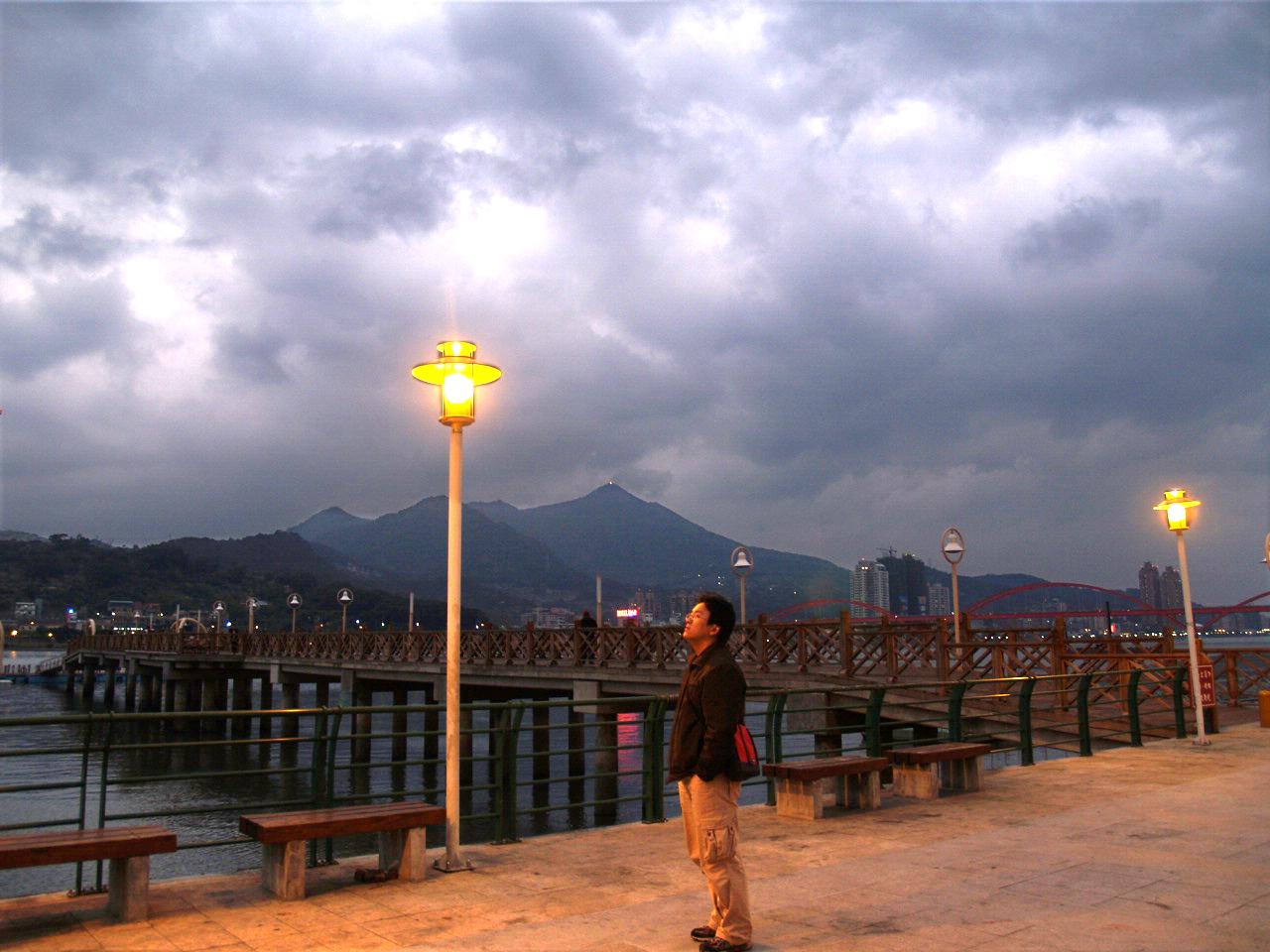}{-0.85cm}{0.4cm}%
		\label{fig:LIME_1}%
	\end{subfigure}\hfill
	\begin{subfigure}{0.122\linewidth}%
		\subcaption{\footnotesize{SRIE~\shortcite{Structure_Li2018}}}%
		\imagewithspy{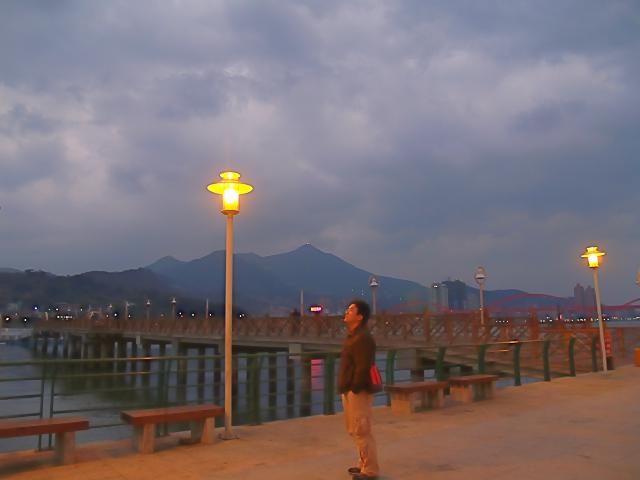}{-0.85cm}{0.4cm}%
		\label{fig:SRIE_1}%
	\end{subfigure}\hfill
	\begin{subfigure}{0.122\linewidth}%
		\subcaption{\footnotesize{MBLLEN~\shortcite{MBLLEN_Lv2018}}}%
		\imagewithspy{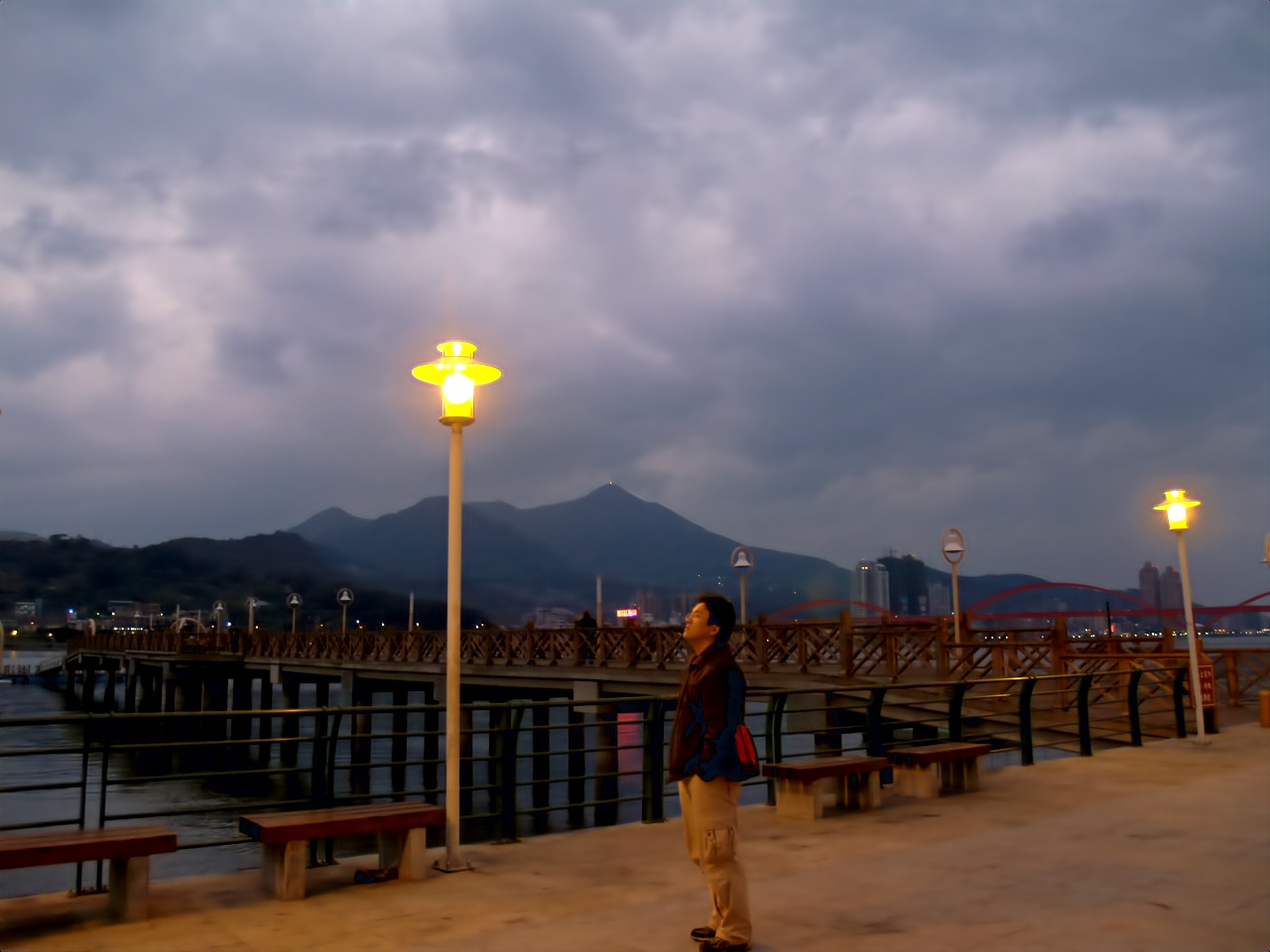}{-0.85cm}{0.4cm}%
		\label{fig:MBLLEN_1}%
	\end{subfigure}\hfill
	\begin{subfigure}{0.122\linewidth}%
		\subcaption{\footnotesize{RetinexNet~\shortcite{Deep_Ret_Wei2018}}}%
		\imagewithspy{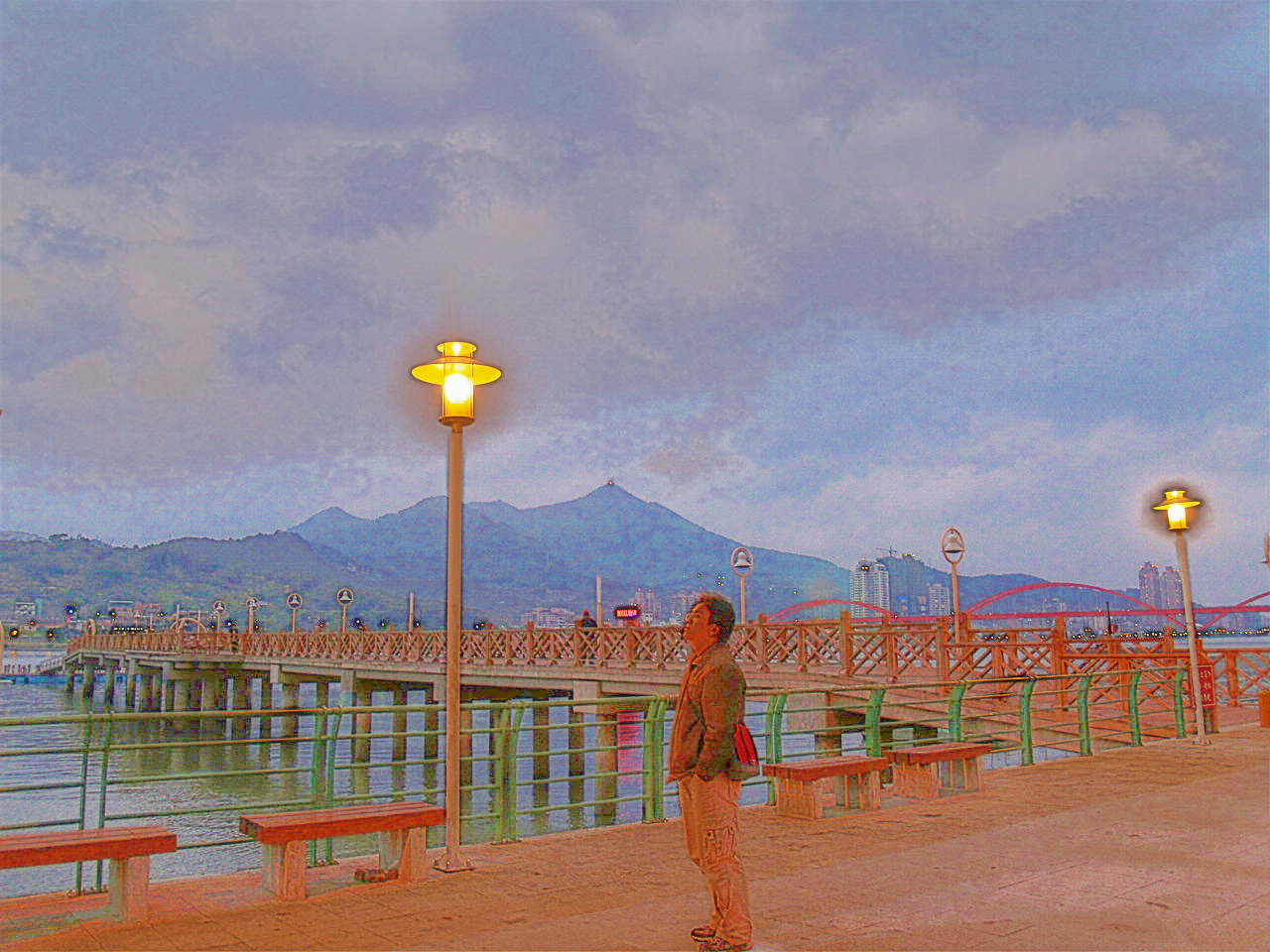}{-0.85cm}{0.4cm}%
		\label{fig:qual_result_images:RetinexNet_1}%
	\end{subfigure}\hfill
	\begin{subfigure}{0.122\linewidth}%
		\subcaption{\footnotesize{Zero-DCE~\shortcite{Zero_DCE_Guo2020}}}%
		\imagewithspy{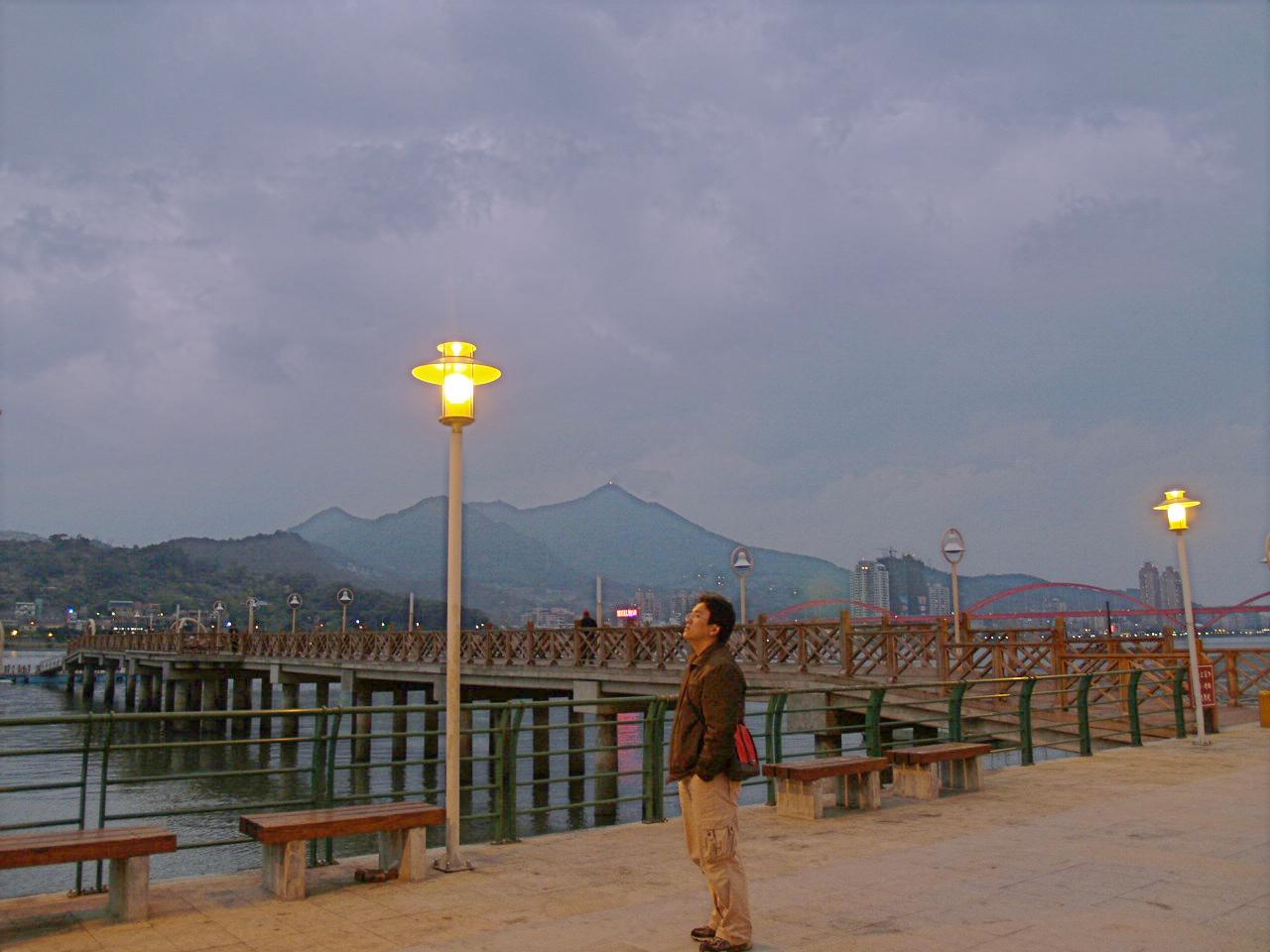}{-0.85cm}{0.4cm}%
		\label{fig:qual_result_images:Zero_DCE_1}%
	\end{subfigure}\hfill
	\begin{subfigure}{0.122\linewidth}%
		\subcaption{\footnotesize{LLVE~\shortcite{Zhang_Temporal2021}}}%
		\imagewithspy{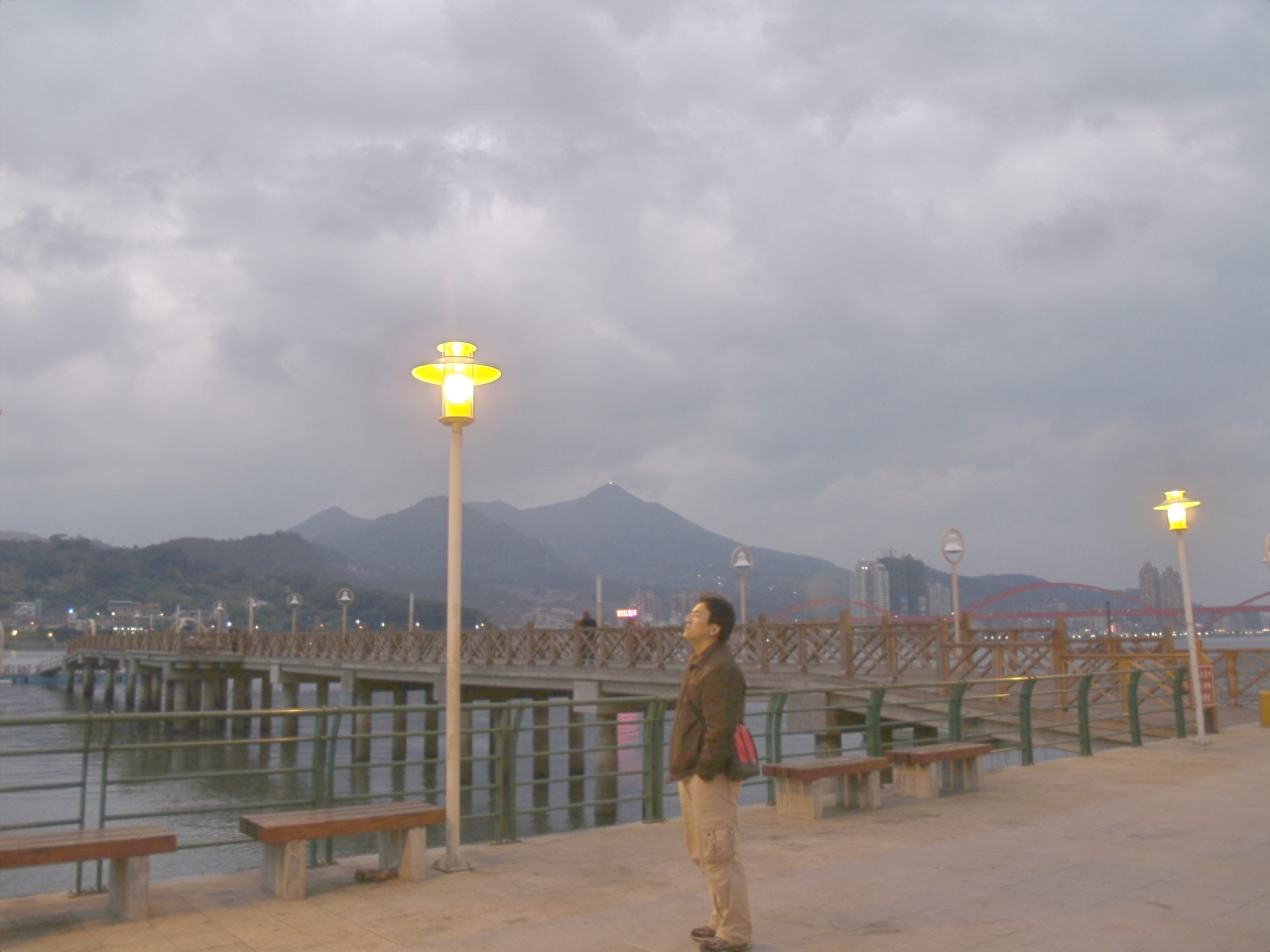}{-0.85cm}{0.4cm}%
		\label{fig:qual_result_images:LLVE_1}%
	\end{subfigure}
	\begin{subfigure}{0.122\linewidth}%
		\subcaption{\footnotesize{Ours}}%
		\imagewithspy{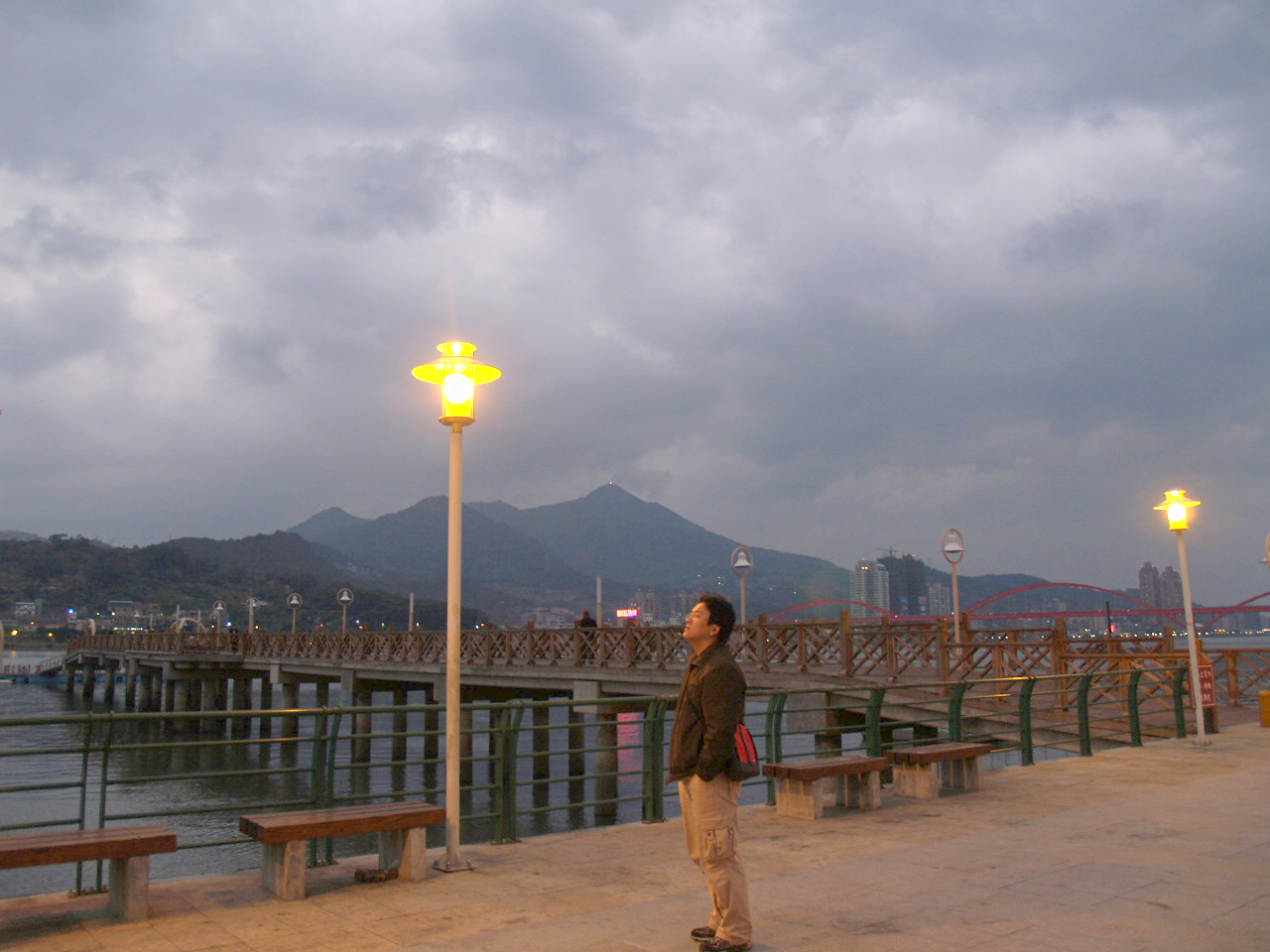}{-0.85cm}{0.4cm}%
		\label{fig:qual_result_images:Ours_1}%
	\end{subfigure}\\[-2.7ex]
\begin{subfigure}{0.122\linewidth}%
	\imagewithspy{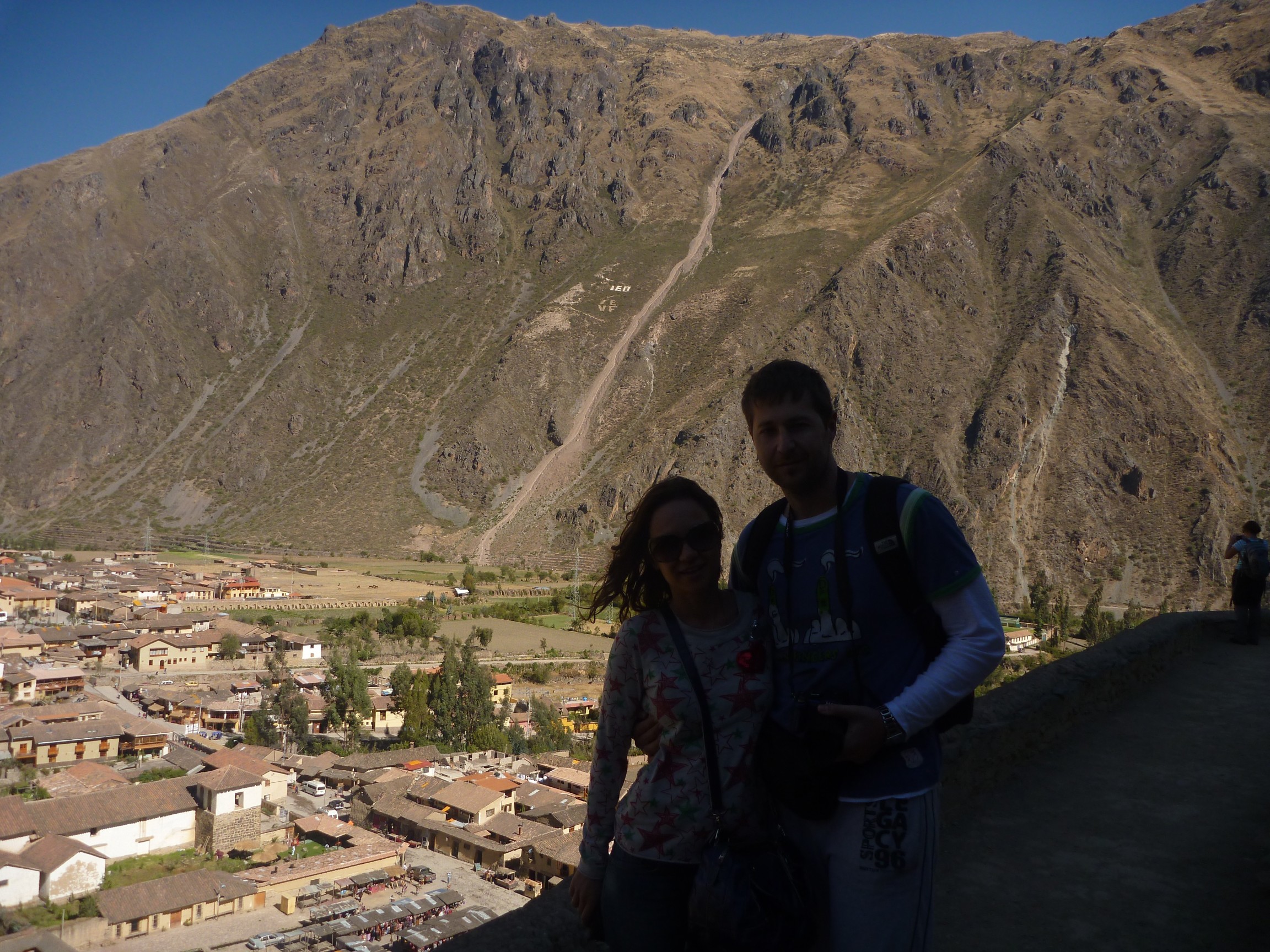}{0.1cm}{-0.12cm}%
	\label{fig:qual_result_images:input_2}%
\end{subfigure}\hfill
\begin{subfigure}{0.122\linewidth}
	\imagewithspy{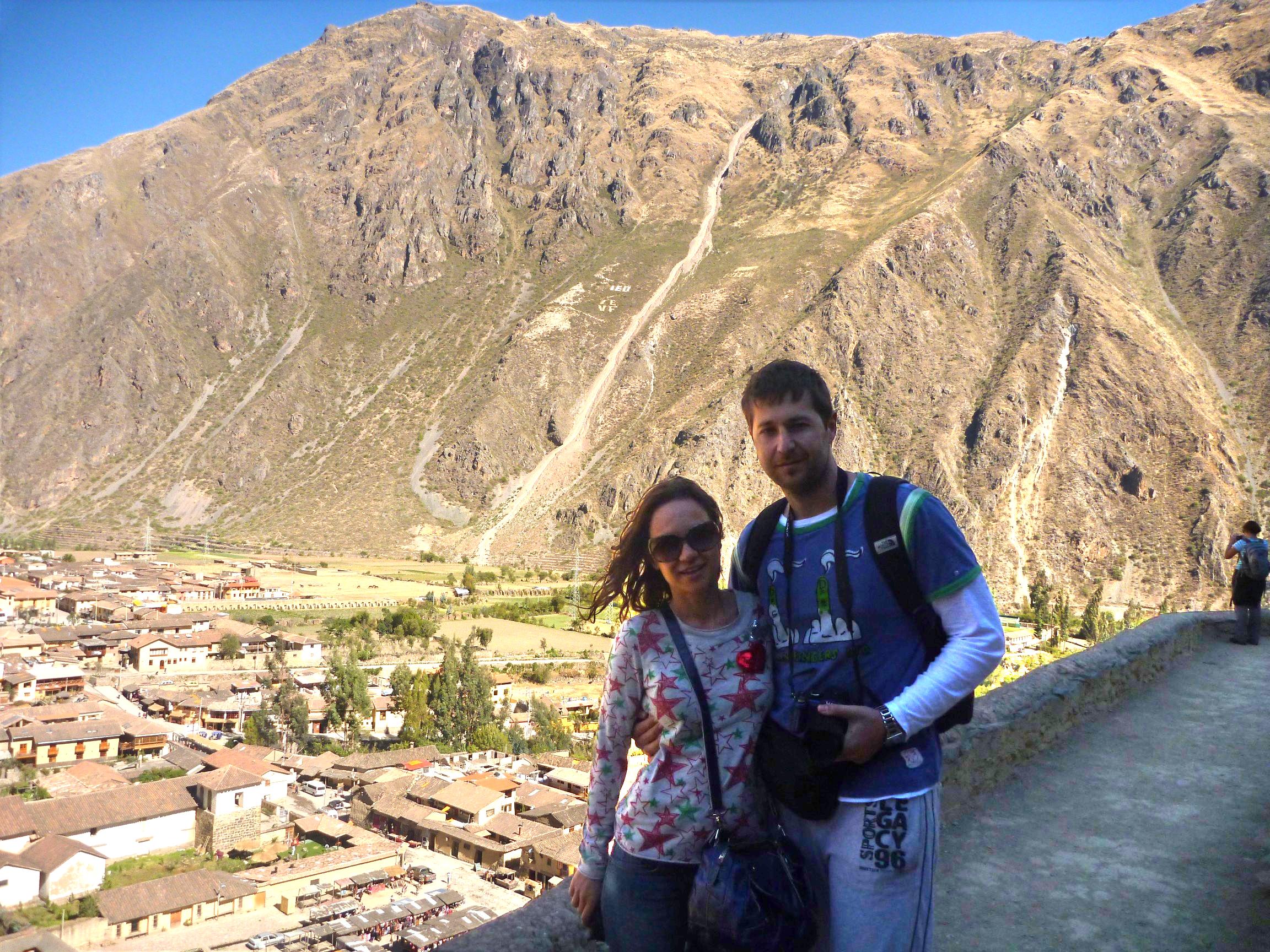}{0.1cm}{-0.12cm}%
	\label{fig:qual_result_images:LIME_2}%
\end{subfigure}\hfill
\begin{subfigure}{0.122\linewidth}%
	\imagewithspy{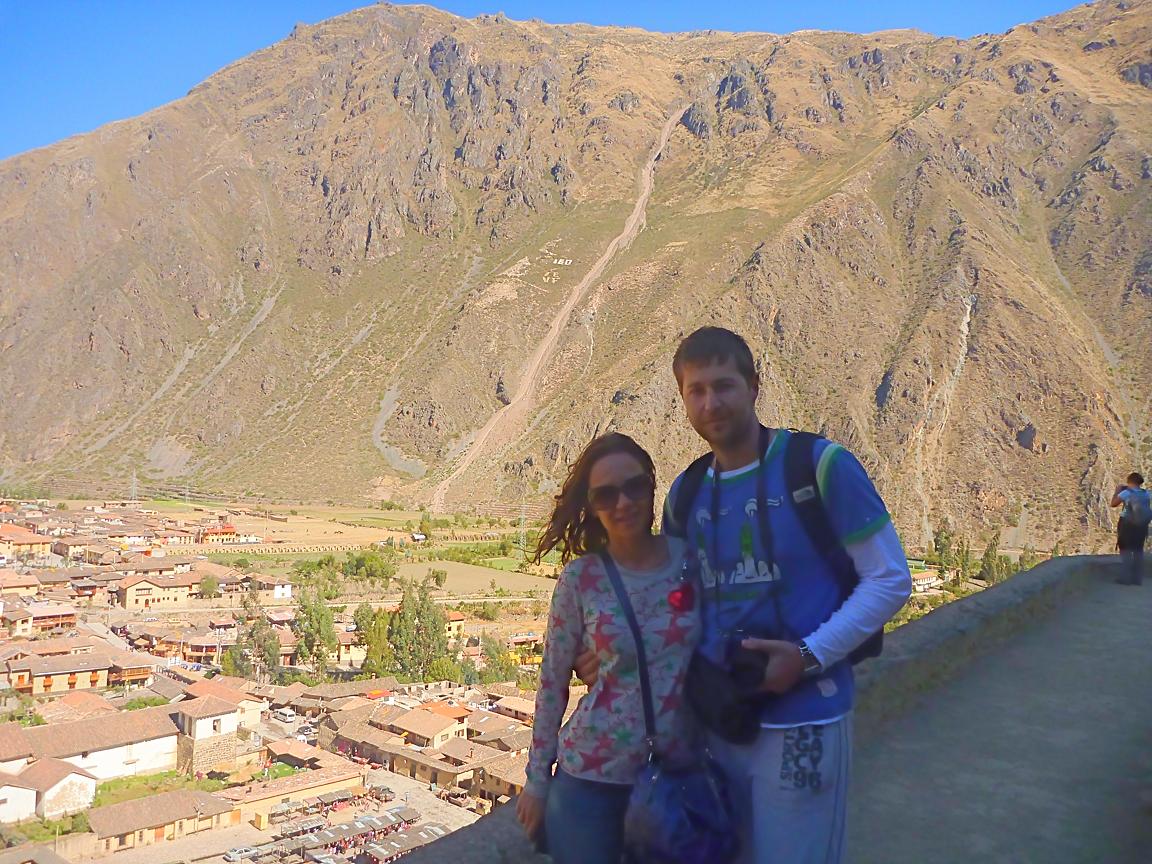}{0.1cm}{-0.12cm}%
	\label{fig:qual_result_images:SRIE_2}%
\end{subfigure}\hfill
\begin{subfigure}{0.122\linewidth}%
	\imagewithspy{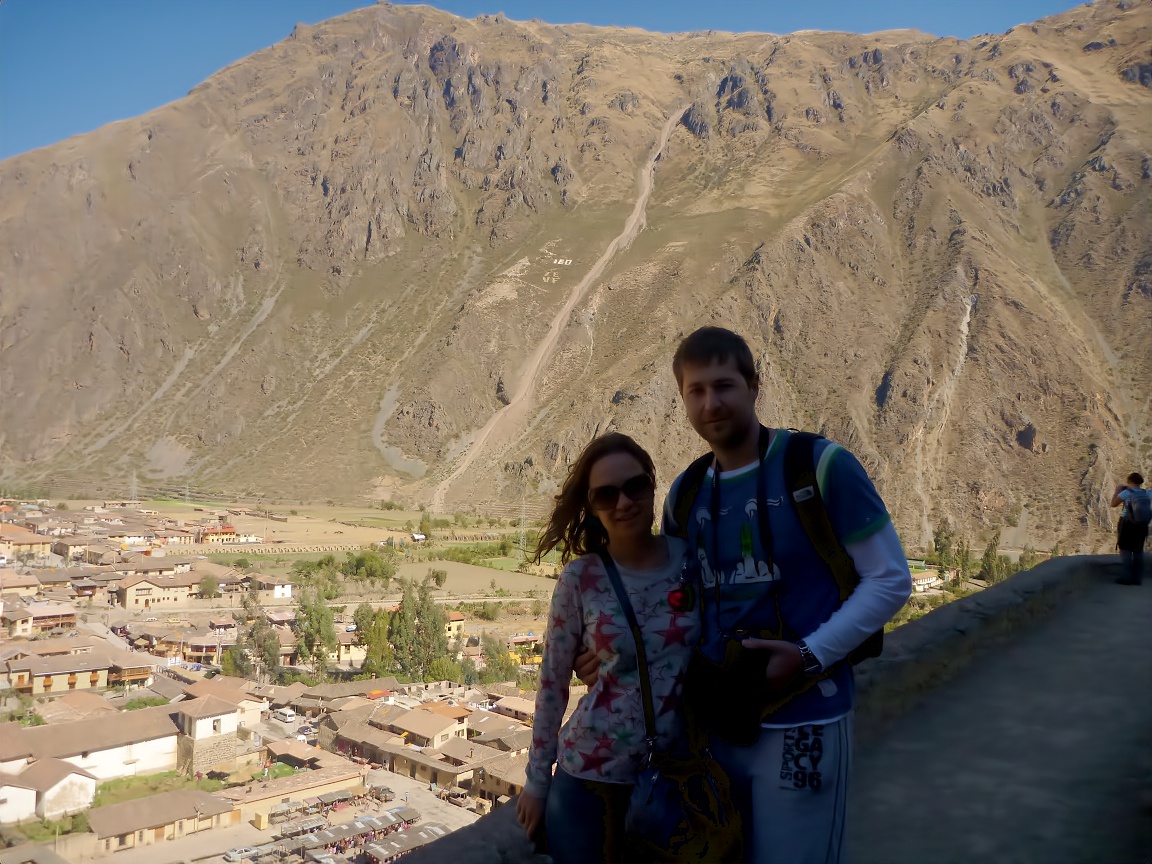}{0.1cm}{-0.12cm}%
	\label{fig:qual_result_images:MBLLEN_2}%
\end{subfigure}\hfill
\begin{subfigure}{0.122\linewidth}%
	\imagewithspy{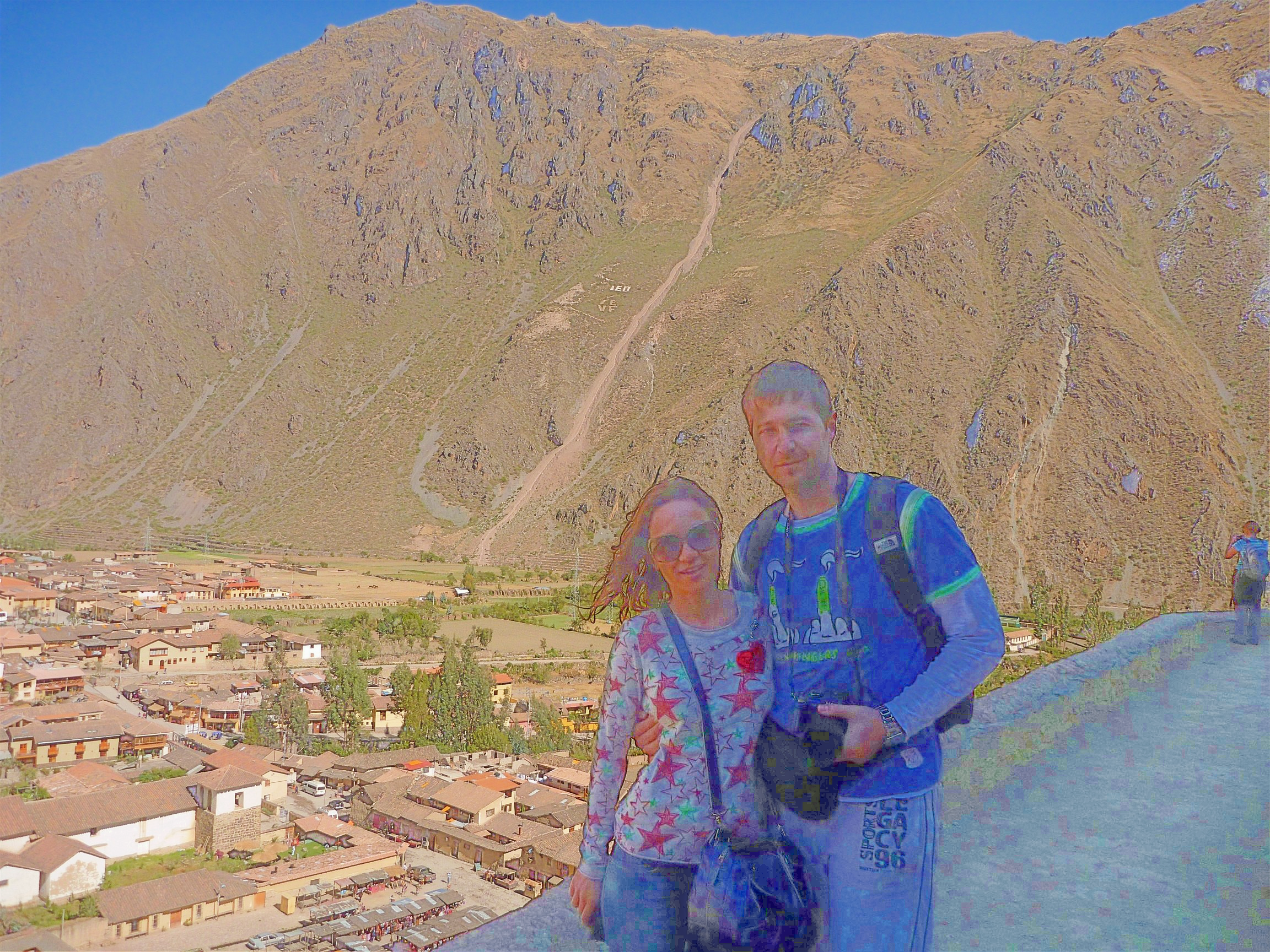}{0.1cm}{-0.12cm}%
	\label{fig:qual_result_images:RetinexNet_2}%
\end{subfigure}\hfill
\begin{subfigure}{0.122\linewidth}%
	\imagewithspy{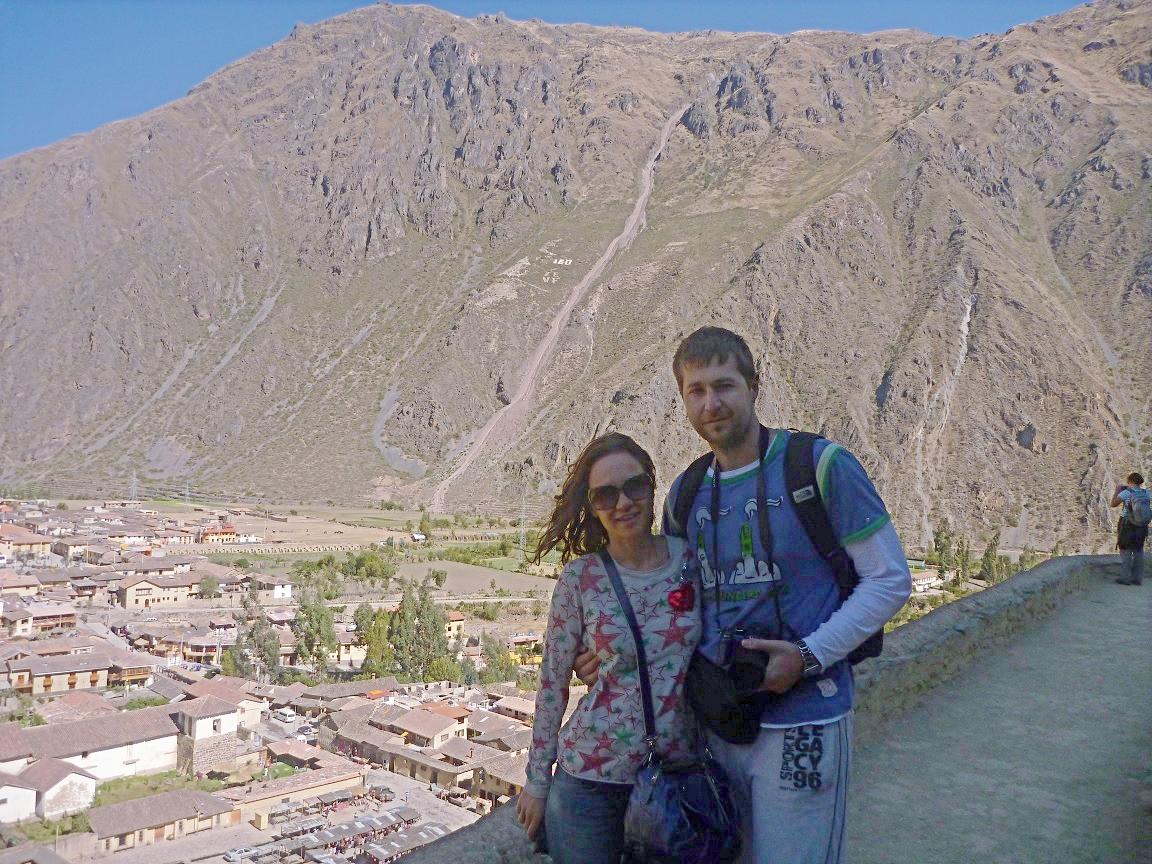}{0.1cm}{-0.12cm}%
	\label{fig:qual_result_images:Zero_DCE_2}%
\end{subfigure}\hfill
\begin{subfigure}{0.122\linewidth}%
	\imagewithspy{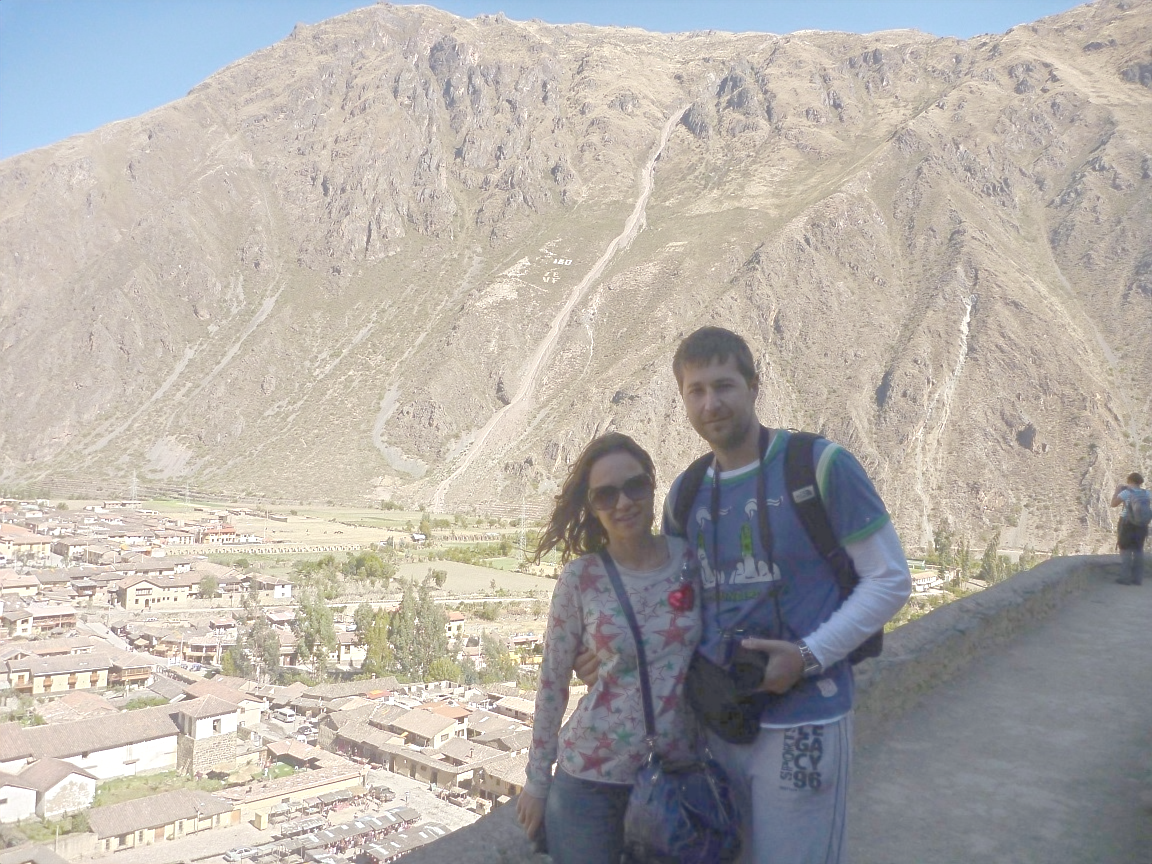}{0.1cm}{-0.12cm}%
	\label{fig:qual_result_images:LLVE_2}%
\end{subfigure}
\begin{subfigure}{0.122\linewidth}%
	\imagewithspy{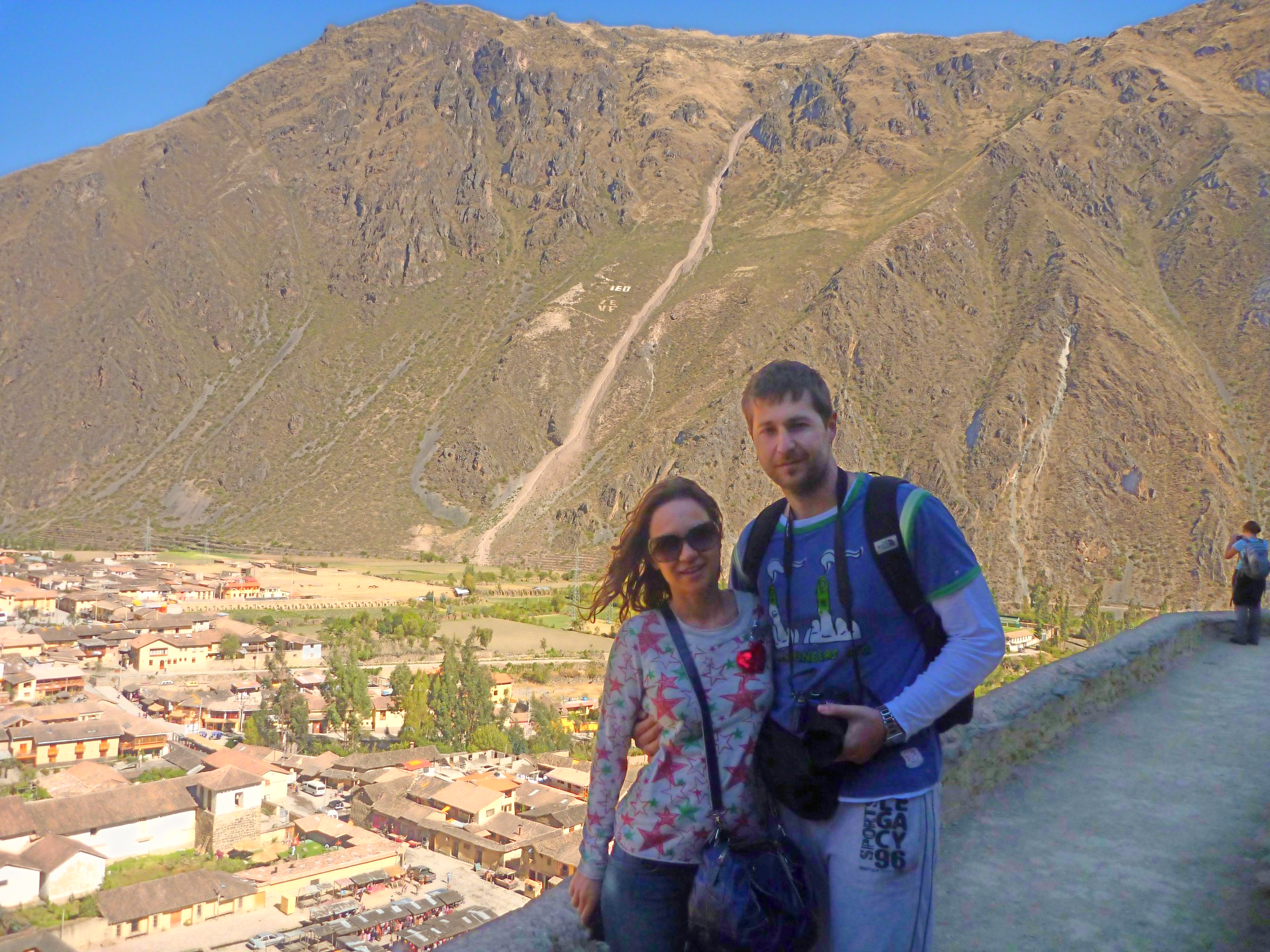}{0.1cm}{-0.12cm}%
	\label{fig:qual_result_images:Ours_2}%
\end{subfigure}\\[-2.7ex]
\begin{subfigure}{0.122\linewidth}%
	\imagewithspy{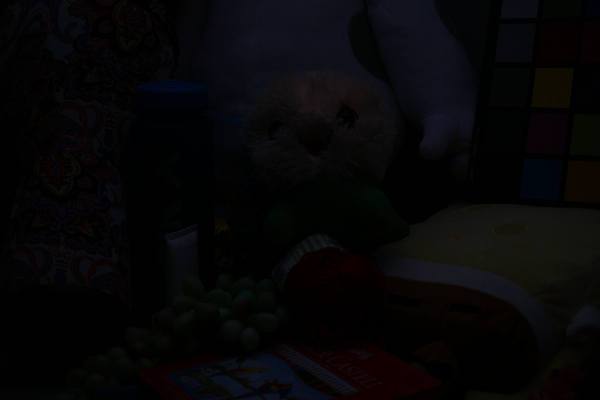}{-0.7cm}{0.3cm}%
	\label{fig:qual_result_images:input_3}%
\end{subfigure}\hfill
\begin{subfigure}{0.122\linewidth}%
	\imagewithspy{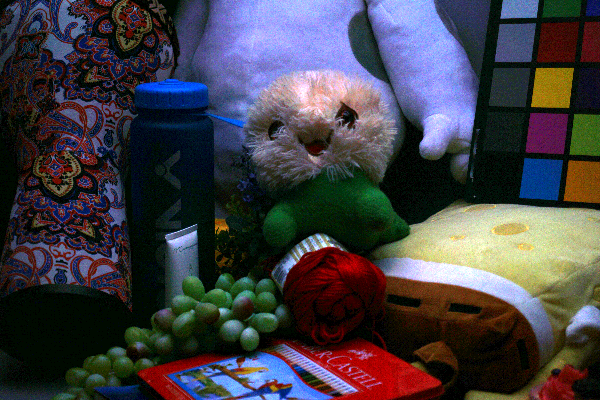}{-0.7cm}{0.3cm}%
	\label{fig:qual_result_images:LIME_3}%
\end{subfigure}\hfill
\begin{subfigure}{0.122\linewidth}%
	\imagewithspy{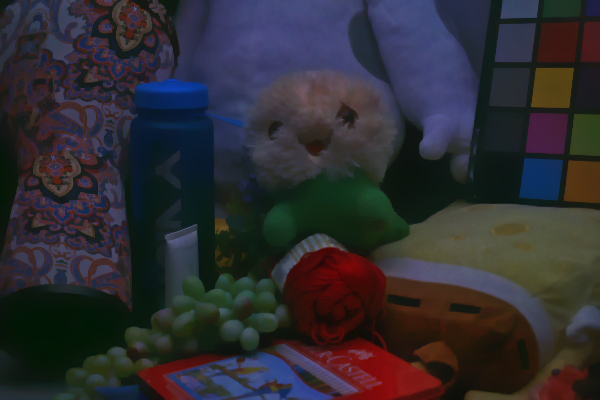}{-0.7cm}{0.3cm}%
	\label{fig:qual_result_images:SRIE_3}%
\end{subfigure}\hfill
\begin{subfigure}{0.122\linewidth}%
	\imagewithspy{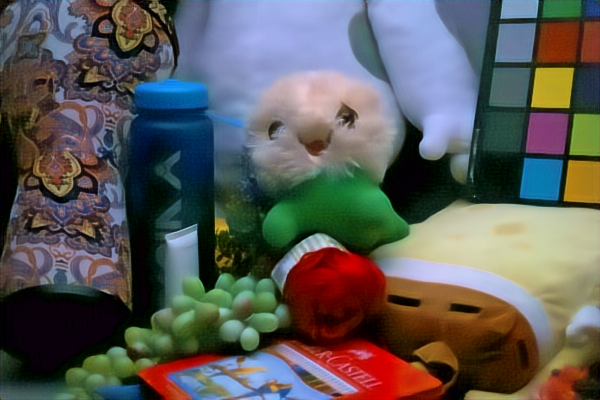}{-0.7cm}{0.3cm}%
	\label{fig:qual_result_images:MBLLEN_3}%
\end{subfigure}\hfill
\begin{subfigure}{0.122\linewidth}%
	\imagewithspy{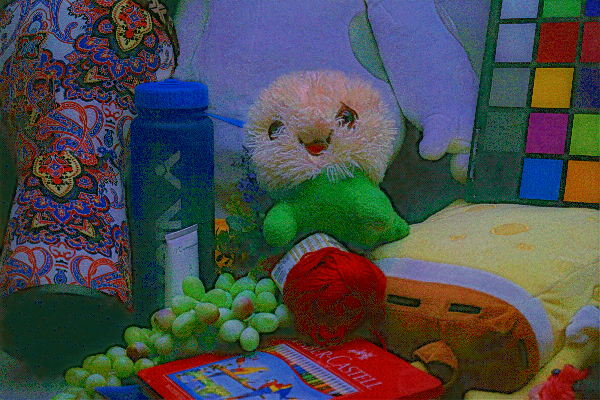}{-0.7cm}{0.3cm}%
	\label{fig:qual_result_images:RetinexNet_3}%
\end{subfigure}\hfill
\begin{subfigure}{0.122\linewidth}%
	\imagewithspy{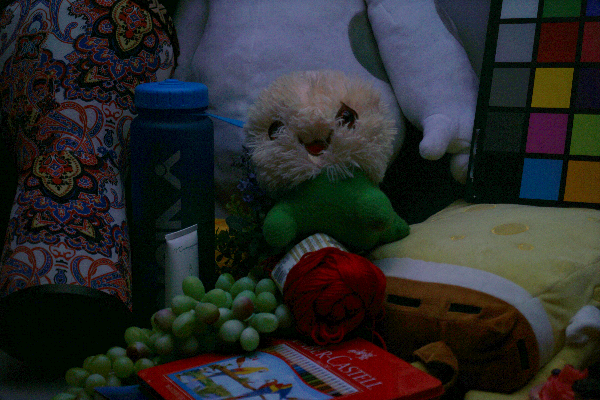}{-0.7cm}{0.3cm}%
	\label{fig:qual_result_images:Zero_DCE_3}%
\end{subfigure}\hfill
\begin{subfigure}{0.122\linewidth}%
	\imagewithspy{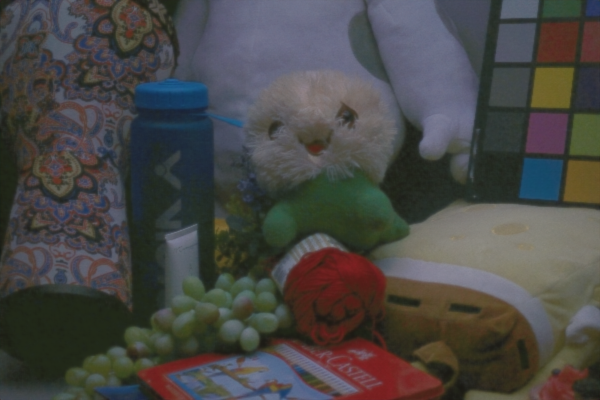}{-0.7cm}{0.3cm}%
	\label{fig:qual_result_images:LLVE_3}%
\end{subfigure}
\begin{subfigure}{0.122\linewidth}%
	\imagewithspy{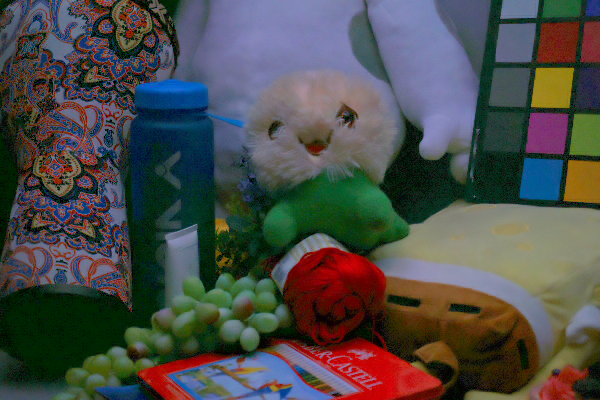}{-0.7cm}{0.3cm}%
	\label{fig:qual_result_images:Ours_3}%
\end{subfigure}\\[-2.7ex]
\begin{subfigure}{0.122\linewidth}%
	\imagewithspy{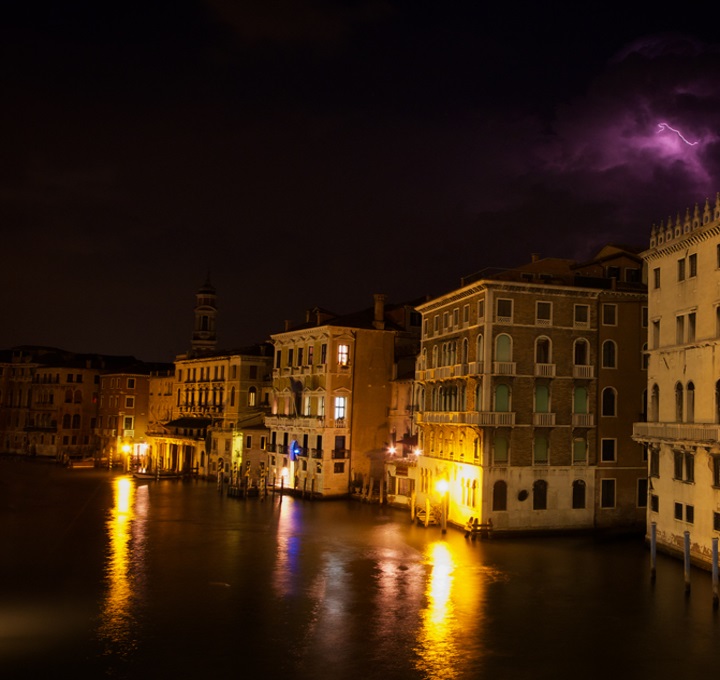}{0.9cm}{0.6cm}%
	\label{fig:qual_result_images:input_4}%
\end{subfigure}\hfill
\begin{subfigure}{0.122\linewidth}%
	\imagewithspy{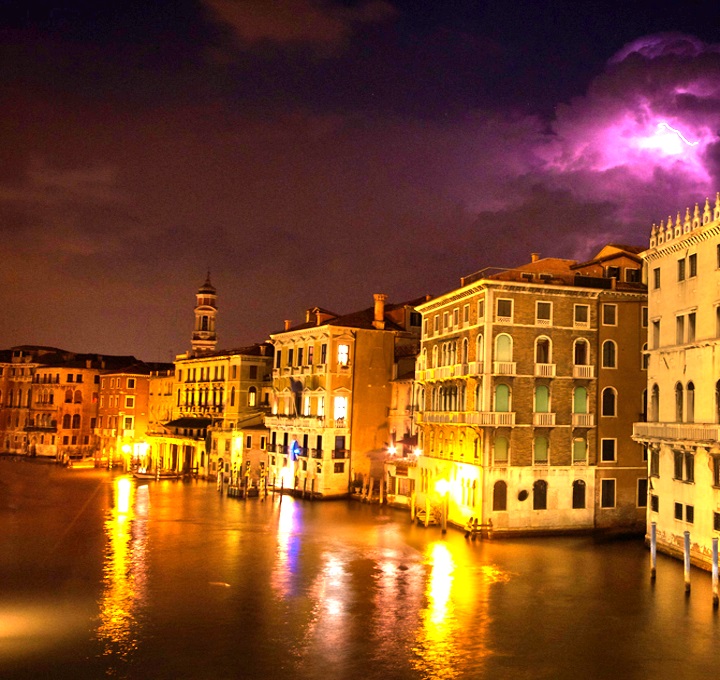}{0.9cm}{0.6cm}%
	\label{fig:qual_result_images:LIME_4}%
\end{subfigure}\hfill
\begin{subfigure}{0.122\linewidth}%
	\imagewithspy{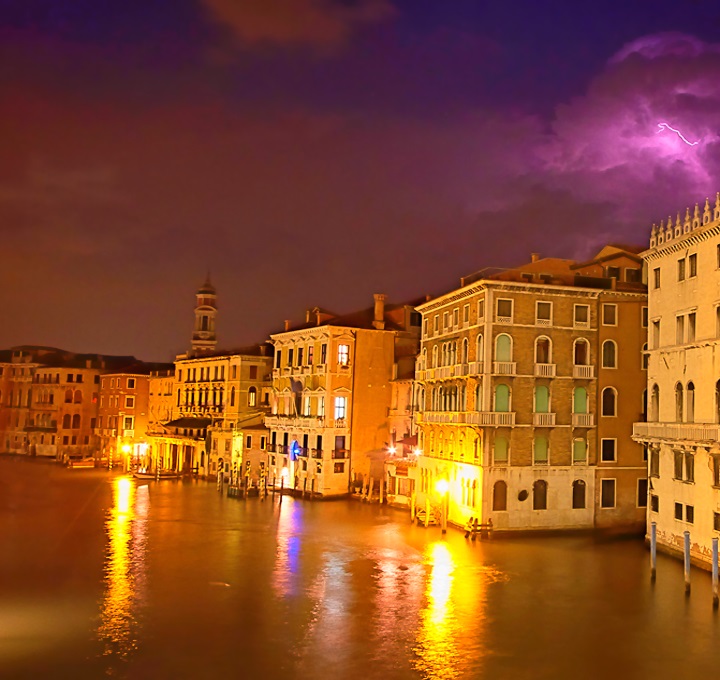}{0.9cm}{0.6cm}%
	\label{fig:qual_result_images:SRIE_4}%
\end{subfigure}\hfill
\begin{subfigure}{0.122\linewidth}%
	\imagewithspy{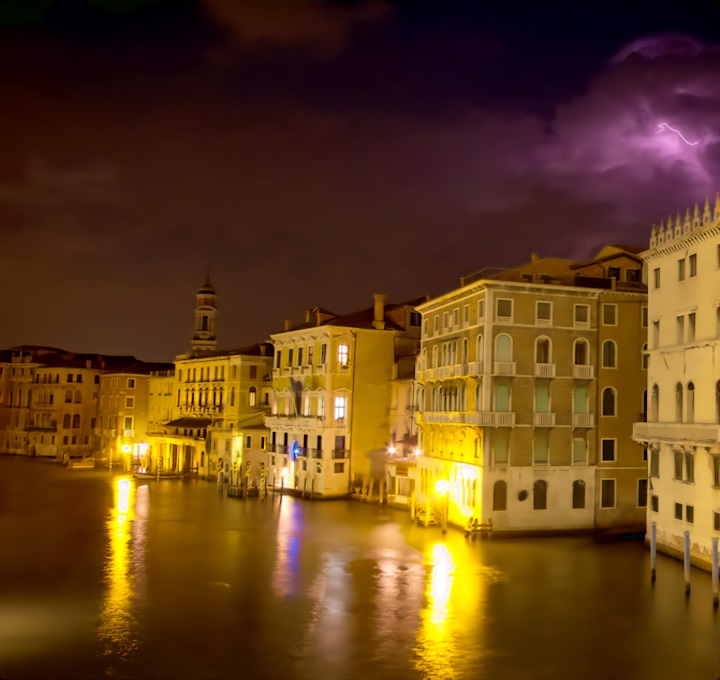}{0.9cm}{0.6cm}%
	\label{fig:qual_result_images:MBLLEN_4}%
\end{subfigure}\hfill
\begin{subfigure}{0.122\linewidth}%
	\imagewithspy{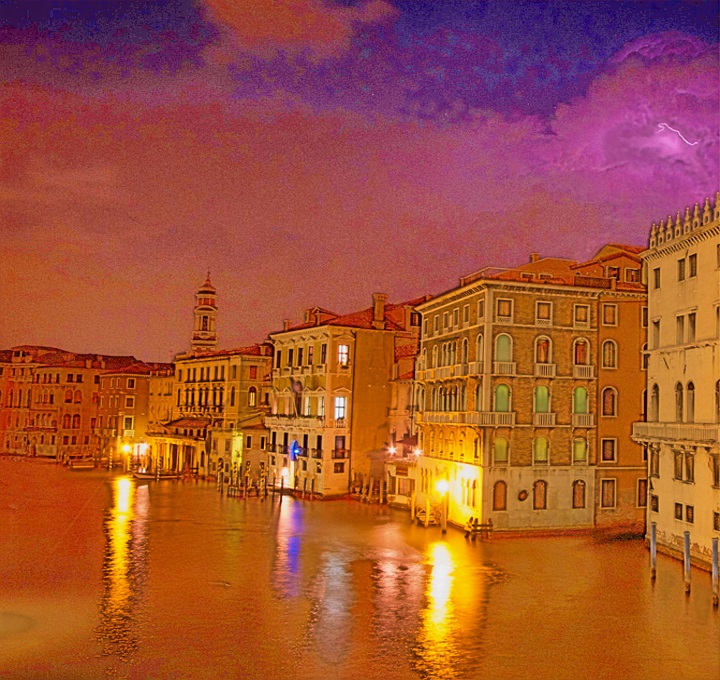}{0.9cm}{0.6cm}%
	\label{fig:qual_result_images:RetinexNet_4}%
\end{subfigure}\hfill
\begin{subfigure}{0.122\linewidth}%
	\imagewithspy{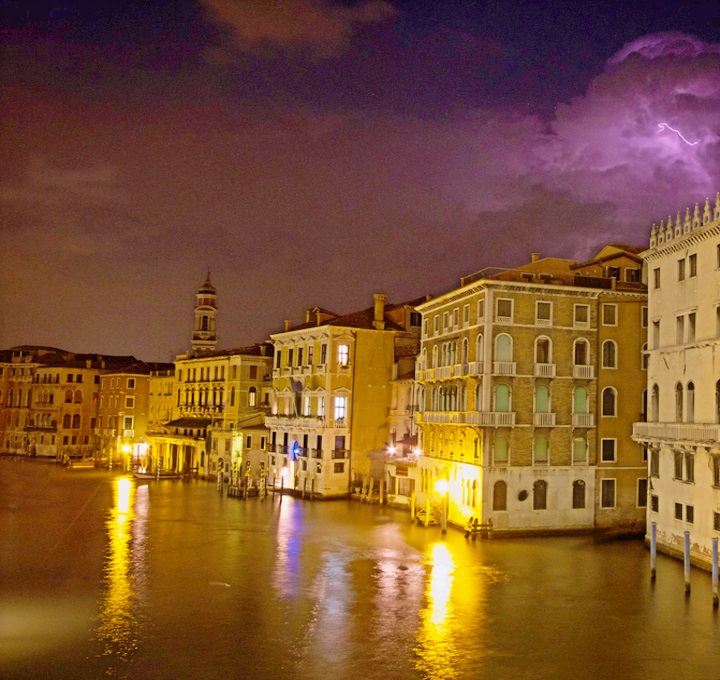}{0.9cm}{0.6cm}%
	\label{fig:qual_result_images:Zero_DCE_4}%
\end{subfigure}\hfill
\begin{subfigure}{0.122\linewidth}%
	\imagewithspy{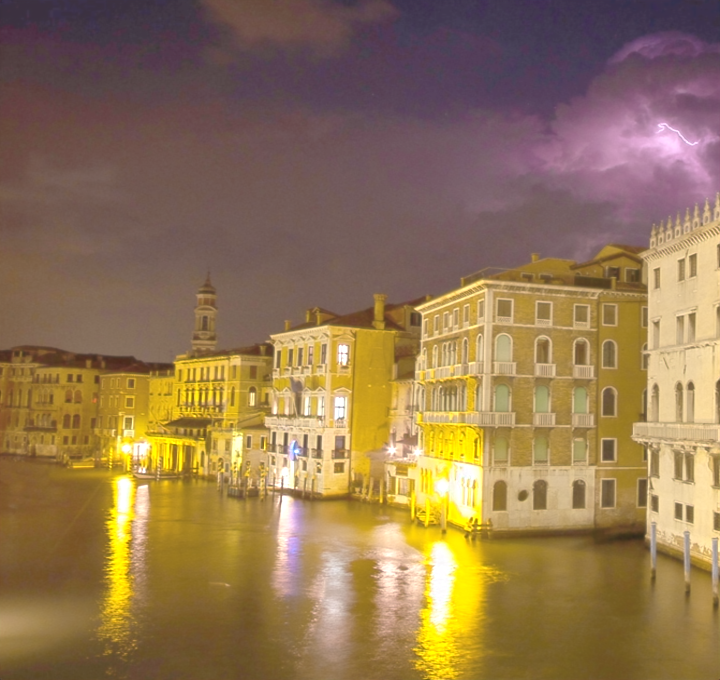}{0.9cm}{0.6cm}%
	\label{fig:qual_result_images:LLVE_4}%
\end{subfigure}\hfill
\begin{subfigure}{0.122\linewidth}
	\imagewithspy{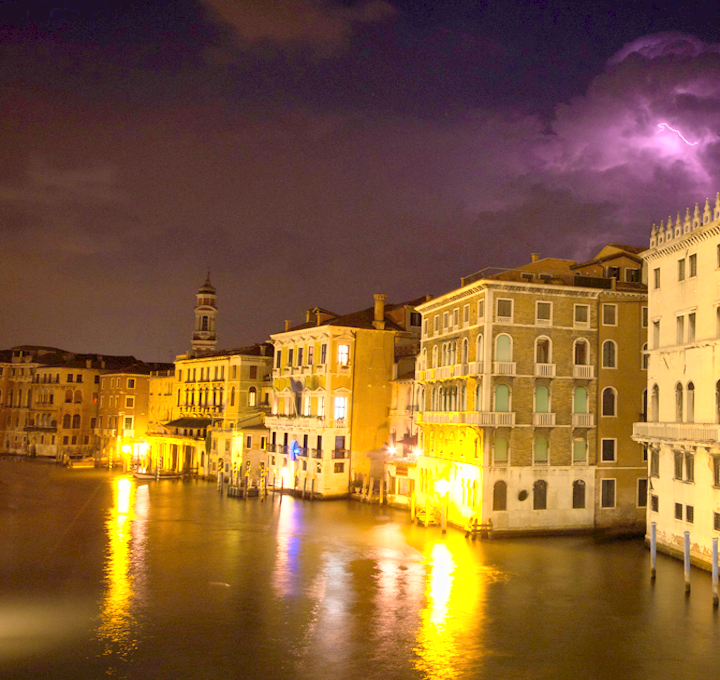}{0.9cm}{0.6cm}%
	\label{fig:qual_result_images:Ours_4}%
\end{subfigure}
\caption{Low-light image enhancement results. Input images are taken from LIME~\shortcite{LIME_Guo2017}, DICM~\shortcite{Contrast_Lee2013}, VV~\shortcite{VV_2022}, and LOL~\shortcite{Deep_Ret_Wei2018} datasets.}%
\label{fig:qual_result_images}
\end{figure*}

\subsection{Face Detection in the Dark}
\label{RL:SubSec:FaceDetection}

We investigate the performance of low-light enhancement methods for increasing the face-detection accuracy on low-light images. 
Specifically, following the settings presented in Li \etal\shortcite{Survey_Li2021}, we use 500 randomly sampled images from the DARK FACE dataset \shortcite{poor_visibility_benchmark} to measure performance of the state-of-the-art \ac{DSFD} \shortcite{li2019dsfd} trained on the WIDER FACE dataset \shortcite{yang2016wider}. 
We use the author's \ac{DSFD} implementation \shortcite{githubli2019dsfd} with a non-maximum suppression threshold of $0.3$ and evaluate using the dark face UG2 challenge evaluation tool \shortcite{github_darkfaceeval}. 
\Cref{fig:facedetection_PR_curve} depicts the precision-recall curves as well as average precision (AP) under a 0.5 IoU threshold. 
The results show that all low-light enhancement methods achieve a significant improvement in precision and recall over the unprocessed images. 
Overall, both our method variants outperform all other methods, with the exception of a precision threshold above 0.85, where RetinexNet~\shortcite{Deep_Ret_Wei2018} has marginally better precision-recall rates. 
Our best performing variant uses a simple \ac{AC} adjustment without denoising or exposure fusion, indicating that more sophisticated methods may smooth or otherwise discard high-frequency information important for face detection.

\subsection{Run-time Performance Evaluation}
\label{RL:SubSec:Quantitative}

All our experiments were performed on an average PC using Microsoft Windows 10 as operating system, with a 2.2 GHz (Intel i7) CPU, 16 GB of RAM, and a Nvidia GTX 1050 Ti graphics card with 4 GB VRAM.
Our full algorithm, implemented with C++ and CUDA (v10.0), runs at real-time for VGA resolution images (\Cref{tab:run_time_eval}) and at interactive frame rates on HD and FHD resolution images. 
Unlike ours, most of the existing techniques are either not able to handle QHD resolution or are very slow for the given hardware configuration. 
Excluding \ac{DAC}, our full version performs better than all the other methods except Zero-DCE~\shortcite{Zero_DCE_Guo2020}.
While \ac{AC} forms the basis of our approach, more than $90\%$ of the processing time is spent on multi-pyramid based exposure fusion. 
If we simply denoise the \ac{AC}, the result thus obtained has artifacts in the form of over-exposedness and lack of details however is already comparable to existing approaches (\Cref{fig:dac_results}).
The \ac{DAC}, our fast variant, can thus potentially serve as a preview of the enhanced output and for further interactive parameter editing. \vspace{-0.5ex}

\begin{table}[]
\caption{Run-time performance of various methods in milliseconds. The top three run-time performance values for each resolution are shown in \textcolor{red}{red}, \textcolor{blue}{blue}, and \textcolor{brown}{brown} colors respectively. For FHD and QHD resolution certain methods do not work for the given hardware configuration due to lack of memory throwing Out-of-Memory (OOM) exception. Note, that LIME and SRIE make use of only CPU and is based on MATLAB code while the other learning-based methods make use of GPU and are based on Tensorflow/Pytorch.}
\label{tab:run_time_eval}
\begin{tabular}{|l|c|c|c|c|}
	\hline
	\small{Meth. \textbackslash Res.}     & \begin{tabular}[c]{@{}c@{}}VGA\\ \small{640 $\times$ 480}\end{tabular} & \begin{tabular}[c]{@{}c@{}}HD\\ \small{1280 $\times$ 720}\end{tabular} & \begin{tabular}[c]{@{}c@{}}FHD\\ \small{1920 $\times$ 1080}\end{tabular} & \begin{tabular}[c]{@{}c@{}}QHD\\ \small{2560 $\times$ 1440}\end{tabular} \\ \hline
	LIME       & 0.579e3                                                 & 1.942e3                                                 & 6.449e3                                                   &  \textcolor{brown} {\textbf{10.178e3}}                                                  \\ \hline
	SRIE       & 11.816e3                                                & 49.834e3                                                & OOM                                                                & OOM                                                                \\ \hline
	MBLLEN     &  0.426e3                          &1.298e3                         &3.014e3                            & OOM                                                                \\ \hline
	RetinexNet & 1.031e3                                                 & 3.714e3                                                 & 7.586e3                                                   & 17.542e3                                                  \\ \hline
	LLVE       & 0.107e3                          & 0.312e3                                                 & 0.699e3                                                   & OOM                                                                \\ \hline
	Zero-DCE   & \textcolor{red} {\textbf{4.685}}                                                          & \textcolor{red} {\textbf{11.771}}                                                         & \textcolor{red} {\textbf{25.753}}                                                           & OOM                                                                \\ \hline
	Ours (DAC) & \textcolor{blue} {\textbf{5.389}}                                   & \textcolor{blue} {\textbf{13.934}}                                  & \textcolor{blue} {\textbf{26.125}}                                    & \textcolor{red} {\textbf{51.503}}                                                           \\ \hline
	Ours (Full)       & \textcolor{brown} {\textbf{61.408}}                                                         &  \textcolor{brown} {\textbf{195.758}}                                                        &  \textcolor{brown} {\textbf{432.808}}                                                          & \textcolor{blue} {\textbf{732.705}}                                                          \\ \hline
\end{tabular}
\end{table}

%% file: alive_discussion.tex
\section{Discussion}
\label{RL:Sec:Discussion}

Most of the existing methods, including ours, face three major challenges for \ac{LLIE}. First is the trade-off between under- and over-exposedness. 
In order to expose the low-lit regions within an image, one might over-expose existing well-exposed parts. 
We approached the above to a large degree by making use of an exposure sequence and multi-pyramid based blending. 
As a generic approach, one can compute the degree of exposure for different image regions, as an exposure mask, in a pre-processing step and use it for further processing. 
Second is the introduction and amplification of noise while enhancing images.
To remove this noise, we use \ac{NLM} denoising that provides plausible results. 
However, improved and efficient denoising technique specially tailored for noises in low-lit images will give better results.
Thirdly, the enhancement process can result in changes in perceived color. 
For us, such change is limited due to counter-balancing effect of $\alpha$ and $\gamma$ on the perceived colorfulness (see supplementary material). 

\noindent \textbf{Limitation}: Among the above challenges we are least effective in terms of noise-removal as we employ a moderate denoising scheme for the sake of better run-time performance and handling of high-resolution images. 
Further, for certain images we might require careful fine tuning of parameters for a better trade-off.

%% file: alive_conclusion.tex
\section{Conclusions and Future Work}
\label{RL:Sec:Conclusions}

This paper presents a simple yet efficient technique to enhance low-light images and videos. 
The key to our approach is Adaptive Chromaticity that allows to increase the image brightness in a straightforward manner. 
The \ac{DAC} is already comparable to state-of-the-art methods and can be potentially used for a fast enhancement preview.
To further improve results, we generate a virtual exposure sequence by computing multiple adaptive chromaticities for the given low-light image followed by a multi-pyramid based fusion.   
Experimental results validate the advancement of our approach in comparison to various state-of-the-art alternatives.
For the above, we perform both quantitative and qualitative evaluation including a subjective user study. 
We believe that our approach can be used to improve the visual quality of low-light images for further processing.
As part of future work we would like to improve the denoising step of our algorithm and potentially use the multi-scale nature of exposure-fusion for this purpose.
For videos we would like to use the neighboring frames to improve the denoising as well as enhancement quality.